\documentclass[10pt,twocolumn,letterpaper]{article}

\usepackage{iccv}
\usepackage{times}
\usepackage{epsfig}
\usepackage{graphicx}
\usepackage{amsmath}
\usepackage{amssymb}
\usepackage{array}
\usepackage{caption}
\usepackage{subcaption}
\usepackage{multirow}

\usepackage{microtype}

\usepackage[pagebackref=true,breaklinks=true,letterpaper=true,colorlinks,bookmarks=false]{hyperref}

\iccvfinalcopy % *** Uncomment this line for the final submission

\def\iccvPaperID{667} % *** Enter the ICCV Paper ID here

\newcommand{\cov}[0]{\text{var}}

\ificcvfinal\fi

\begin{document}

%%%%%%%%% TITLE
%\title{Leveraging computer-generated imagery for deep feature analysis}
%\title{Understanding deep features with synthetic imagery}
\title{Understanding deep features with computer-generated imagery}

\author{Mathieu Aubry\\
\'{E}cole des Ponts ParisTech \ \ \ UC Berkeley\\
{\tt\small mathieu.aubry@imagine.enpc.fr}
% For a paper whose authors are all at the same institution,
% omit the following lines up until the closing ``}''.
% Additional authors and addresses can be added with ``\and'',
% just like the second author.
% To save space, use either the email address or home page, not both
\and
Bryan C. Russell\\
Adobe Research\\
{\tt\small brussell@adobe.com}
}

\maketitle

%\thispagestyle{fancy}

%%%%%%%%% ABSTRACT
\begin{abstract}
We introduce an approach for analyzing the variation of features generated  by convolutional neural networks (CNNs) with respect to scene factors that occur in natural images. 
Such factors may include object style, 3D viewpoint, color, and scene lighting configuration.
Our approach analyzes CNN feature responses corresponding to different scene factors by controlling for them via rendering using a large database of 3D CAD models.
The rendered images are presented to a trained CNN and responses for different layers are studied with respect to the input scene factors. 
We perform a decomposition of the responses based on knowledge of the input scene factors and analyze the resulting components. 
In particular, we quantify their relative importance in the CNN responses and visualize them using principal component analysis. We show qualitative and quantitative results of our study on three  CNNs trained on large image datasets: AlexNet~\cite{Krizhevsky12}, Places~\cite{Zhou14}, and Oxford VGG~\cite{Chatfield14}. 
We observe important differences across the networks and CNN layers for  different scene factors and object categories.
Finally, we demonstrate that our analysis based on computer-generated imagery translates to the network representation of natural images.
\end{abstract}

%%%%%%%%% BODY TEXT
\section{Introduction}

The success of convolutional neural networks (CNNs)~\cite{Krizhevsky12,LeCun89} raises fundamental questions on how their learned representations encode variations in visual data.
For example, how are different layers in a deep network influenced by different scene factors, the task for which the network was trained for, or the choice in network architecture?
%Given the success of convolutional neural networks (CNNs)~\cite{Krizhevsky12,LeCun89}, the question of the image representation they learn became crucial. 
%How does the representation depends from variations in the data? %How does it depend 
%From the task the network was trained for? From the network architecture? From the network layer considered? 
% What is the nature of the learned representation, particularly with respect to the large datasets on which they are trained? \mathieu{ -> I am not sure what ``the nature of the representation mean''} 
These questions are important as CNNs with different architectures and trained/fine tuned for different tasks have shown to perform differently~\cite{Karayev14,Zhou14} or have different feature response characteristics~\cite{Zhou15}.
An analysis of the features may help with understanding the tradeoffs across different trained networks and may inform the design of new architectures. It may also help the choice of CNN features for tasks where training or fine tuning a network is not possible, e.g.\ due to lack of labeled data.
%Convolutional neural networks (CNNs)~\cite{LeCun89}, coupled with large labeled image databases (e.g.\ ImageNet~\cite{ILSVRCarxiv14} and Places~\cite{Zhou14}), dropout regularization~\cite{Krizhevsky12}, and efficient implementations on modern GPU hardware, have yielded state-of-the-art results for several recognition tasks, such as object classification~\cite{Chatfield14,Krizhevsky12,Szegedy14}, object detection~\cite{Girshick14}, and scene classification~\cite{Zhou14}.
%Given the success of CNNs, the question arises: What is the nature of the learned representation, particularly with respect to the large datasets on which they are trained?
%This question is important as CNNs trained or fine tuned for different tasks have shown to perform differently~\cite{Karayev14,Zhou14} or have different feature response characteristics~\cite{Zhou15}.
%Such an analysis of the features may help with understanding the tradeoffs across different trained networks and may inform the design of new architectures.
%Moreover, a number of applications have used the output of a given layer as a feature (e.g.\ AlexNet~\cite{Krizhevsky12} ``fc7'' layer), so understanding the nature of its representational capacity may have practical benefit. 

Prior work has focused on a part-based analysis of the learned convolutional filters.
Examples include associating filters with input image patches having
maximal response~\cite{Girshick14}, deconvolution starting from a given filter response~\cite{Zeiler14}, 
%Prior work has visualized the learned convolutional filters by associating input image patches with 
%maximal response for a given filter~\cite{Girshick14}, via deconvolution starting from a given filter response~\cite{Zeiler14}, 
or by masking the input to recover the receptive field of a given filter~\cite{Zhou15} to generate ``simplified images''~\cite{Biederman95,Tanaka93}.
Such visualizations typically reveal the parts of an object~\cite{Zeiler14} (e.g.\ ``eye'' of a cat) or scene~\cite{Zhou15} (e.g.\ ``toilet'' in bathroom).
While these visualizations reveal the nature of learned filters, they largely ignore the question of the dependence of the CNN representation on continuous factors that may influence the depicted scene, such as 3D viewpoint, scene lighting configuration, and object style.%the variety of other factors that comprise a depicted scene, such as 3D viewpoint, scene lighting configuration, and object style.

In this paper, we study systematically how different scene factors that arise in natural images are represented in a trained CNN. 
Example factors may include those intrinsic to an object or scene, such as category, style, and color, and extrinsic ones, such as 3D viewpoint and scene lighting configuration.
%\mathieu{I don' t understand the beginning of this paragraph+ I think we should give a special place to object style, for which we are really the first ones (not possible with NORB)}
Studying the variations associated with such factors is a nontrivial task as it requires (i) input data where the factors can be independently controlled and (ii) a procedure for detecting, visualizing, and quantifying each factor in a trained CNN.% We address both issues.

%To address the former issue, one could collect a database of natural images with the different factors controlled, e.g.\ NORB~\cite{Lecun04} and RGB-D object~\cite{Lai11} datasets, where different objects are rotated on a turntable and lighting is varied during image capture.
%However, such datasets are currently relatively difficult and costly to collect, particularly at scale.

To overcome the challenges associated with obtaining input data, we leverage computer-generated (CG) imagery to study trained CNNs.
CG images offer several benefits.
First, there are stores of 3D content online (e.g.\ Trimble 3D Warehouse), with ongoing efforts to curate and organize the data for research purposes (e.g. ModelNet~\cite{Wu15}).
Such data spans many different object categories and styles. 
Moreover, in generating CG images we have control over all rendering parameters, which allows us to systematically and densely sample images for any given factor. 
A database of natural images captured in controlled conditions and spanning different factors of variations, e.g. the NORB~\cite{Lecun04}, ETH-80~\cite{leibe2003} and RGB-D object~\cite{Lai11} datasets, where different objects are rotated on a turntable and lighting is varied during image capture, are difficult and costly to collect.
Moreover they do not offer the same variety of object styles present in 3D model collections, nor the flexibility given by rendering.

Given a set of rendered images generated by varying one or more factors, we analyze the responses of a layer for a trained CNN (e.g.\ ``pool5'' of AlexNet~\cite{Krizhevsky12}).
%We design an informative decomposition of the features, where most terms depend on a single factor. We study the relative importance of each factor in the representation and compute its principal modes of variation using principal component analysis (PCA) .
%For a single independent factor, we perform principal component analysis (PCA) to visualize and quantify the principal modes of variation in the feature space with respect to the factor.
%We also consider the case when multiple factors are present.
We perform a decomposition of the responses based on knowledge of the input scene factors, which allows us to quantify the relative importance of each factor in the representation.
Moreover, we visualize the responses via principal component analysis~(PCA).
%We visualize and quantify the responses via principal component analysis (PCA).
%Moreover, when multiple scene factors are present we analyze a linear decomposition of the responses based on knowledge of the input scene factors.
%Our decomposition allows us to quantify the relative importance of each factor in the representation.
\\[-7mm]

\paragraph{Contributions.}
Our technical contribution is an analysis of contemporary CNNs  trained on large-scale image datasets via computer generated images, particularly rendered from 3D models.
From our study we observe:
\begin{itemize}
\item Features computed from image collections that vary along two different factors, such as style and viewpoint, can often be approximated by a linear combination of features corresponding to the factors, especially in the higher layers of a CNN.
\item Sensitivity to viewpoint decreases progressively in the last layers of the CNNs.
Moreover, the VGG fc7 layer appears to be less sensitive to viewpoint than AlexNet and Places.
\item Relative to object style, color is more important for the Places CNN than for AlexNet and VGG. 
This difference is more pronounced for the background color than for foreground.
\item The analysis we perform on deep features extracted from rendered views of 3D models is related to understanding their representation in natural images.
%\item The representation of rendered views from 3D models is pertinent for understanding their representation in natural images.
%\item Finally, we show in section~\ref{sec:natural_images} that the CG images can be realistic enough to transfer the analysis to natural images.%, as we show.% how our study with CG images translates to natural images).
\end{itemize}
%We plan to release all our data and code.

%
%Describe high-level conclusions of experiments.
%\begin{itemize}
%\item invariance learned by vgg to geometric variations is more important, and allows the features to focus on the object appearence. This may be related a diminution of the ``style'' dimension between pool5 and fc6 that doesn't exist for AlexNet
%\item the importance given to the BG color is much more important for the place CNN than for AlexNet and VGG and increase with the layers. This is also true for the FG color but not as strong. AlexNet and VGG give much more importance to the FG color than the BG color
%\item we visualize how complex transformations (such as translations or 3D rotations) are encoded, and that they differ between categories
%\item the invariance to rendering parameters augments consistently with the layers
%\item it is possible to identify dimension that de-correlate pose/style... The shape (viewpoint and style)  information is even separated enough in the CNN from all other factors (position, scale, texture...) that direct 2d-3d matching with NN makes sense (I am not sure if the dimensionality reduction is important or not, that's tricky to test because of memory issues, but I will try to do it)
%\end{itemize}

\subsection{Related work}
In addition to the prior work to visualize learned CNN filters~\cite{Berkes06,Erhan09,Girshick14,Simonyan13,Zeiler14,Zhou15}, there has been work to visualize hand-designed~\cite{Vondrick2013hoggles} and deep~\cite{Mahendran15} features.
Also related are recent work to understand the quantitative tradeoffs across different CNN layers for networks trained on large image databases~\cite{Agrawal14,Yosinski14} and designing CNN layers manually~\cite{Bruna13}.

Our use of a large CAD model dataset can be seen in the context of leveraging such data for computer vision tasks, e.g.\ object detection~\cite{Aubry14b}.
%A contemporary approach has adapted such data to CNNs to render chair images given style and pose~\cite{Dosovitskiy15}.
%Deep convolutional inverse graphics network~\cite{Kulkarni15}.
%CG2Real~\cite{Johnson11}
Contemporary approaches have used synthetic data with CNNs to render images for particular scene factors, e.g.\ style, pose, lighting~\cite{Dosovitskiy15,Kulkarni15}.

Our PCA feature analysis is related to prior work on studying visual embeddings.
The classic Eigenfaces paper~\cite{Turk01} performed PCA on faces.
Later work studied nonlinear embeddings, such as LLE~\cite{Roweis00} and IsoMap~\cite{Tenenbaum00b}.
Most related to us is the study of nonlinear CNN feature embeddings with the NORB dataset~\cite{Hadsell06}.
In contrast we study large-scale, contemporary CNNs trained on large image datasets.
Finally, our multiple factor study is related to intrinsic image decomposition~\cite{Barrow78}. We note a contemporary approach for separating style and content via autoencoders~\cite{Cheung15}.

\subsection{Overview}

Our deep feature analysis begins by rendering a set of stimuli images by varying one or more scene factors.
%Our deep feature analysis begins by rendering a set of stimuli images with one or more scene factors present.
We present the stimuli images to a trained CNN as input and record the feature responses for a desired layer.
Given the feature responses for the stimuli images we analyze the principal modes of variation in the feature space via PCA (section~\ref{sec:single_factor}).
When more than one factor is present we linearly decompose the feature space with respect to the factors and perform PCA on the feature decomposition (section~\ref{sec:multiple_factors}).
We give details of our experimental setup in section~\ref{sec:experimental_setup} and show qualitative and quantitative results over a variety of synthetic and natural images in section~\ref{sec:results}.

\section{Approach for deep feature analysis}
\label{sec:theorie}

In this section we describe our approach for analyzing the image representation learned by a CNN. % trained on natural images. 
We seek to study how the higher levels of the CNN encodes the diversity present in a set of images. The minimal input for our analysis is a set of related images.
We first describe our approach for analyzing jointly their features.
What we can learn with such an approach is however limited since it cannot identify the origin of the variations of the input images. 
The factors of variation can be, e.g., variations in the style of an object, changes in its position, scaling, 3D rotation, lighting, or color.  For this reason, we then focus on the case when the images are computer generated and we have full control of the different factors. In this case, we seek to separate the influence of the different factors on the representation, analyze them separately, and compare their relative importance.%\bryan{cite reference here?} 
%where a factor is defined as a family of parameters 
%as it undergoes different transformations, 
%These factor of variation can be for example
%To achieve this analysis, we take as input to a set of computer generated images, parameterized by one or more factors we seek to study.
%Given the rendered images we compute their features, seek to identify the influence of each factor, characterize their intrinsic dimensionality and interpret the principal modes of variation.
%, and then how we separate and compare the influence of different factors.

%different factors of variation, such as 

%Motivation: What are the big questions that we want to investigate with our analysis?  We will study: translation, scale, 3D rotation, object texture, scene lighting, color.

\subsection{Image collection analysis}
\label{sec:single_factor} 

We seek to characterize how a CNN encodes a collection of related images, $\Omega$, e.g.\ images depicting a ``car'' or a black rectangle on white background. %be a collection of related images
We sample images $r_\theta\in\Omega$ indexed by $\theta\in\Theta$. %Each image $r_\theta\in\Omega$ is indexed by $\theta\in\Theta$.
%For our analysis, we suppose we have access to  a subset of $\Omega$ that we can index by $\theta\in\Theta$. 
% and is %an independent factor of an object or scene, where
%We wish to characterize a set of factors of a rendered scene.  
%the factor may correspond to, e.g., object style, 3D rotation, scene lighting configuration, object category, or object color.
%We index this image collection by. 
In the case of natural images $\Theta$ is an integer index set over the collection $\Omega$.
In the case of computer-generated images $\Theta$ is a set of parameters corresponding to a scene factor we wish to study (e.g.\ azimuth and elevation angles for 3D viewpoint, 3D model instances for object style, or position of the object in the image for 2D translation). %. We call $r(\theta)$ the image indexed by $\theta$.
%For particular parameters $\theta\in\Theta$, we denote $r(\theta)$ as a rendered image of the scene. 
Given a trained CNN, let $\tilde{F}^L(r_\theta)$ be a column vector of CNN responses for layer $L$ (e.g.\ ``pool5'', ``fc6'', or ``fc7'' in AlexNet~\cite{Krizhevsky12}) to the input image $r_\theta$.

The CNN responses $\tilde{F}^L$ are high-dimensional feature vectors that represent the image information.
However, since $\Omega$ contains related images, we expect the features to be close to each other and their intrinsic dimension to be smaller than the actual feature dimension.
%We seek to identify the principal modes of variation of the features $\tilde{F}^L$.
For this reason, we use principal component analysis (PCA)~\cite{Pearson1901} to identify a set of orthonormal basis vectors that capture the principal modes of variation of the features.

Given centered features $F^L(\theta) = \tilde{F}^L(r_\theta) - \frac{1}{|\Theta|} \sum_{t\in\Theta} \tilde{F}^L(r_t)$, where $|X|$ is the number of elements in set $X$, 
%
%We start by centering the set of features $\tilde{F}^L$:
%
%\begin{equation}
%F^L(\theta) = \tilde{F}^L(r(\theta)) - 
%\frac{1}{|\Theta|} \sum_{t\in\Theta} \tilde{F}^L(r(t))
%\end{equation}
%
%\noindent
%and then compute the eigenvectors associated to the largest eigen-values of the covariance matrix of the features, $\sum_{\theta\in\Theta} F^L(\theta)(F^L(\theta))^T$. 
we compute the eigenvectors associated with the largest eigenvalues of the covariance matrix $\frac{1}{|\Theta|}\sum_{\theta\in\Theta} F^L(\theta)(F^L(\theta))^T$. 
%\bryan{Discuss computational complexity of computing the covariance matrix and alternative methods.  Also mention this in the CNN section to motivate why we study pool5, fc6, and fc7.}
The projection of the features onto the subspace defined by the $D$ components with maximal eigenvalues corresponds to an optimal D-dimensional linear approximation of the features.
We evaluate the intrinsic dimensionality of the features by computing the number of dimensions necessary to explain 95\% of the variance. 
Moreover, we can visualize the embedding of the images by projecting onto the  components with high variance. 
%Moreover, we can use this analysis to visualize the embedding of the images learned by the CNN, using 2D projections of the features on planes defined by components with high variance. 
%\mathieu{I am not very happy with this paragraph, maybe we need to introduce or refer to some probabilistic PCA.}
%Moreover, we can analyze the intrinsic dimensionality of the factor via the total variance of the basis directions with maximal variance.

%This first K components of the PCA can be interpreted as a linear approximation approach can be interpreted as identifying a local linear approximation of the manifold of features of the images we are interested in.
%\paragraph{Dimensionality:}
% $F_L(r_i(\theta))$, $X_{\theta,i}$, $V_{\theta}$, $M_{i}$ and are typically very high dimensional vectors that are able to represent the information included in any kind of images. However, one can reasonably expect their intrinsic dimension to be much smaller, especially if considering only related images. 
%We estimate the intrinsic dimensionality using PCA.

\subsection{Multiple factor analysis}
\label{sec:multiple_factors} 
In the case when we have control of the variation parameters, we can go further and attempt to decompose the features as a linear combination of uncorrelated components associated to the different factors of variation. 
Features decomposing linearly into different factors would be powerful and allow to perform image transformations directly in feature space and, e.g., to compare easily images taken under different viewpoints.%be especially powerful for applications, allowing to learn to identify the position of an object, its illumination, and its viewpoint from a single feature, but also to perform image transformations simply in feature space.
%In addition to studying the different factors independently, we also consider several factors jointly in our deep feature analysis.

Let $\Theta_1,\dots,\Theta_N$ be sets of parameters for $N$ factors of variation we want to study.
We consider an image of the scene with parameters $\theta = (\theta_1,\dots,\theta_N)$, where $\theta\in\Theta = \Theta_1 \times \dots \times \Theta_N$. We assume the $\theta_k$ are sampled independently.
We define marginal features $F^L_k(\theta_k)$ for scene factor $k$ by marginalizing over the parameters for all factors except $k$:

\begin{eqnarray}
F^L_k(t) &=& \mathbb{E}( F^L(\theta)| \theta_k=t)\\
&=&\frac{|\Theta_k|}{|\Theta|} \sum_{\theta\in\Theta| \theta_k=t} F^L(\theta)
\label{eq:marg}
\end{eqnarray}

%\noindent
Similar to section~\ref{sec:single_factor}, we can study via PCA the principal modes of variation over the marginal features $F^L_k$ for each factor $k$.

Finally, we define a residual feature $\Delta^L(\theta)$, which is the difference of the centered CNN features $F^L(\theta)$  and the sum of all the marginal features $F^L_k(\theta_k)$.
This results in the following decomposition:

%We write the centered CNN features $F^L(\theta)$  as a linear combination of terms each depending on single factors $k$ and a residual:

\begin{equation}
F^L(\theta) = \sum_{k=1}^N F^L_k(\theta_k) + \Delta^L(\theta)
\label{eqn:decomposition}
%\Delta^L(\theta) = F^L(\theta) - \sum_{k=1}^N F^L_k(\theta_k)
\end{equation}

%\noindent
%where $\Delta^L(\theta)$ is the residual feature and $F^L_k(\theta_k)$ marginalizes over the parameters for all factors except $k$:
%
%\begin{eqnarray}
%F^L_k(t) &=& \mathbb{E}( F^L(\theta)| \theta_k=t)\\
%&=&\frac{|\Theta_k|}{|\Theta|} \sum_{\theta\in\Theta| \theta_k=t} F^L(\theta)
%\label{eq:marg}
%\end{eqnarray}

\noindent
Using computer-generated images, we can easily compute this decomposition by rendering the images with all the rendering parameters corresponding to the sum in equation~(\ref{eq:marg}). %\bryan{Can we cite related work here for the above assumptions?} 
Direct computation shows that all the terms in decomposition (\ref{eqn:decomposition}) have zero mean and are uncorrelated. This implies:

\begin{equation}
\cov(F^L) = \sum_{k=1}^N \cov(F^L_k) + \cov(\Delta^L)
\end{equation}

We can thus decompose the variance of the features $F^L$ as the sum of the variances associated to the different factors and a residual.  
When analyzing the decomposed features %of the factors for a given CNN layer 
we report the relative variance $R^L_k = \cov(F^L_k)/\cov(F^L)$. 
We also report the relative variance of the residual $R^L_\Delta = \cov(\Delta^L)/\cov(F^L)$.
A factor's relative variance provides an indication of how much the factor is represented in the CNN layer compared to the others. 
A high value indicates the factor is dominant in the layer and conversely a low value indicates the factor is largely negligible compared to the others.
Moreover, a low value of the residual relative variance indicates the factors are largely separated in the layer.
Note that $R^L_\Delta + \sum_{k=1}^N R^L_k = 1$ and the values of $R^L_k$ and $R^L_\Delta$ do not depend on the relative sampling of the different factors.

%We can thus decompose the variance of the features $F^L$ as the sum of the variances associated to the different factors and a residual.  
%The part of the variance explained by $F^L_k$ indicates the importance of the factor $k$ for the representation, while the part of the variance explained by the residual indicates how well the feature separates the influence of the different factors.
%For example a small residual indicates an uncorrelated representation of the different factors. 
%Similar to section~\ref{sec:single_factor}, we can study via PCA the principal modes of variation over the centered CNN features $F^L$, the marginal features $F^L_k$ for each factor $k$, and the residual feature $\Delta^L$.

%When analyzing the decomposed features %of the factors for a given CNN layer we report the relative variance $R^L_k = \cov(F^L_k)/\cov(F^L)$. 
%We also report the relative variance of the residual $R^L_\Delta = \cov(\Delta^L)/\cov(F^L)$.
%A factor's relative variance provides an indication of how much the factor is represented in the CNN layer compared to the others. 
%A high value indicates the factor is dominant in the layer and conversely a low value indicates the factor is largely negligible compared to the others.
%Moreover, a low value of the residual relative variance indicates the factors are largely separated in the layer.
%Note that $R^L_\Delta + \sum_{k=1}^N R^L_k = 1$ and the value of $R^L_k$ and $R^L_\Delta$ does not depend on the relative sampling of the different factors.

\section{Experimental setup}
\label{sec:experimental_setup}

In this section we describe details of our experimental setup.
In particular we describe the details of the CNN features we extract, our rendering pipeline, and the set of factors we seek to study.

\subsection{CNN features}
%We seek to study the responses of CNNs via a large pool of rendered images.
We study three trained CNN models: AlexNet~\cite{Krizhevsky12}, winner of the 2012 ImageNet Large-Scale Visual Recognition Challenge (ILSVRC)~\cite{ILSVRCarxiv14}, Places~\cite{Zhou14}, which has the same architecture as AlexNet but trained on a large image database depicting scenes, and Oxford VGG~\cite{Chatfield14} CNN-S network. 
In particular, we study the features of the higher layers ``pool5'', ``fc6'', and ``fc7'' of these networks.
Note that the Oxford VGG architecture is different and the dimension of its ``pool5'' layer is two times larger than AlexNet and Places.
We use the publicly-available CNN implementation of Caffe~\cite{jia2014caffe} to extract features for the different layers and pre-trained models for AlexNet, Places, and Oxford VGG from their model zoo.

\subsection{Computer-generated imagery}

We present two types of image stimuli as input: (i) 2D abstract stimuli consisting of constant color images or rectangular patches on constant background, which are described in detail in section~\ref{sec:2d}, and (ii) rendered views from 3D models.
For the latter we seek to render different object categories spanning many different styles from a variety of viewpoints and under different illumination conditions.
We used as input CAD 3D models from the ModelNet database~\cite{Wu15}, which contains many models having different styles for a variety of object classes.
We downloaded the CAD models in Collada file format for the following object classes: chair (1261 models), car (485 models), sofa (701 models), toilet (191 models) and bed (258 models).
%We manually removed noisy models (e.g.\ models with extra objects, such as a person, ground plane, or other objects in the scene), which resulted in \bryan{FILL IN NUMBER OF MODELS} CAD models in total for our study.
%Each CAD model consists of sets of 3D vertices, vertex colors, and triangle indices.  
%For many models there are also a set of image textures and 2D texture coordinates for the vertices.
We adapted the publicly-available OpenGL renderer from \cite{Aubry14}, which renders a textured CAD model with matte surfaces under fixed lighting configuration and allows the viewpoint to be specified by a 3$\times$4 camera matrix~\cite{Hartley04}.
We render the models under different lighting conditions and with different uniform colors.
%In the following sections we will describe how we vary the rendering parameters and viewpoint for the different experiments.

We show results on two categories that have received the most attention in 3D-based image analysis: cars \cite{Glasner11,Pepik2012}, and chairs \cite{Aubry14b,Dosovitskiy15}. Detailed results for three other categories are presented in the supplementary material and our quantitative results are
%, except when mentioned otherwise, 
averages over the 5 categories. 

\begin{figure}[t]
\centering
\begin{subfigure}[b]{0.22\linewidth}%45
       \includegraphics[width=0.98\linewidth]{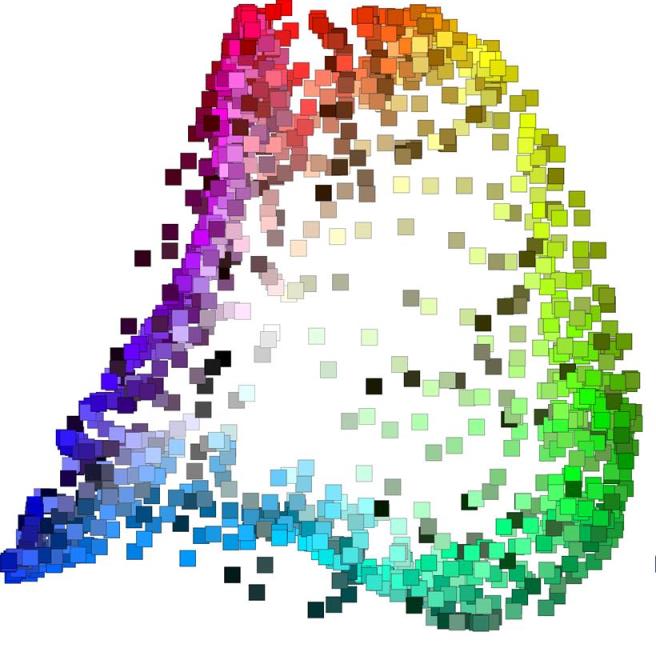}
        \caption{AlexNet\\ dims.\ 1, 2}
    %     \label{fig:colora}
\end{subfigure}
\begin{subfigure}[b]{0.25\linewidth}%5
       \includegraphics[width=0.98\linewidth]{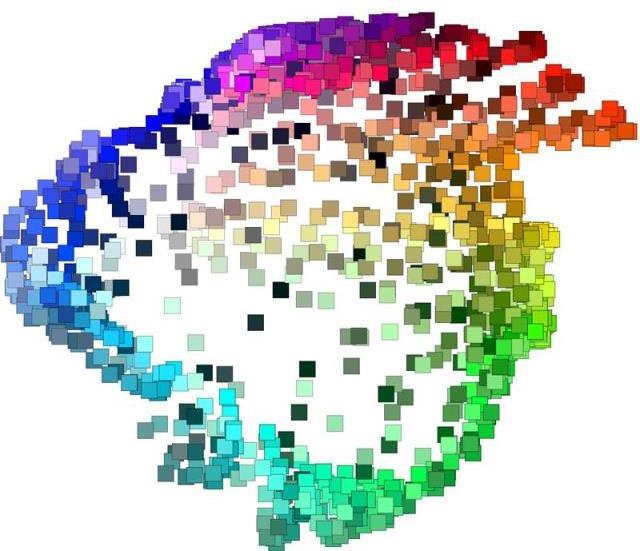}
        \caption{Places\\ dims.\ 1, 2}
       %  \label{fig:colora}
\end{subfigure}%\\
\begin{subfigure}[b]{0.26\linewidth}%5
       \includegraphics[width=0.98\linewidth]{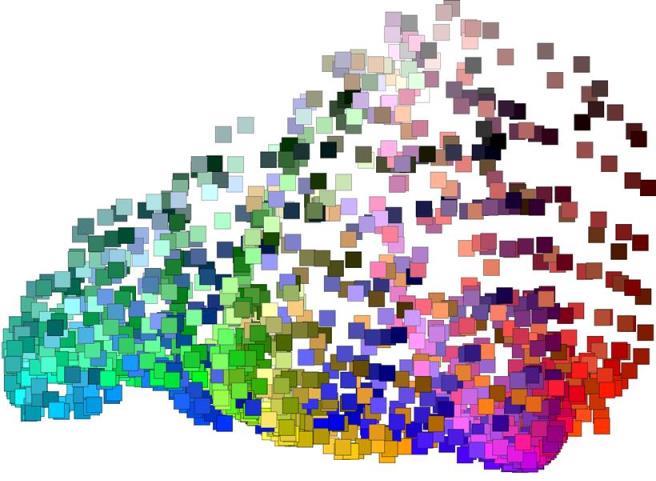}
        \caption{VGG\\ dims.\ 1, 2}
       %  \label{fig:colora}
\end{subfigure}%
\begin{subfigure}[b]{0.23\linewidth}%45
       \includegraphics[width=0.98\linewidth]{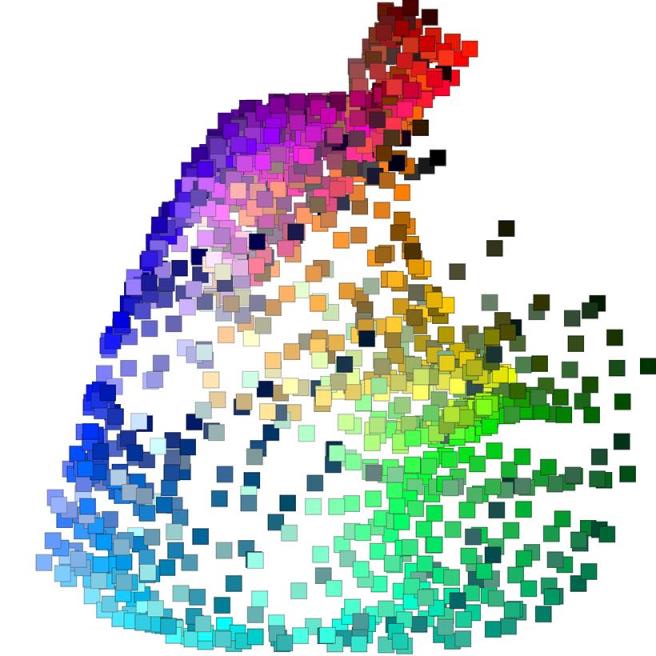}
        \caption{VGG\\  dims.\ 2, 3}
       %  \label{fig:colora}
\end{subfigure}%
\caption{
PCA embeddings of constant-color images for the fc7 layer of different CNNs.
The AlexNet and Places embeddings are similar to a hue color wheel, with more variation visible for the green and blue channels.
}
\label{fig:color}
\end{figure}

\subsection{3D scene factors} 
\label{sec:3D_factors}
We study two types of factors affecting the appearance of a scene: (i) intrinsic factors -- object category, style, and color, and (ii) extrinsic factors -- 2D position, 2D scale, 3D viewpoint, and scene lighting configuration.
The object category and style factors are specified by the CAD models from ModelNet.
For object color we study grayscale matte surfaces (specified by Lambertian surface model), and constant-colored matte surfaces with the color uniformly sampled on a grid in RGB colorspace.  %\bryan{Provide more details on the constant color experiment.  How many images for each factor?}
For the 2D extrinsic factors, we uniformly sample 2D positions along a grid in the image plane and vary the 2D scale linearly.
For 3D viewpoint we manually aligned all the 3D models to a canonical coordinate frame, i.e.\ all models are consistently oriented with respect to gravity and face the same direction, orbit the object at constant 3D distance and uniformly sample the azimuth and elevation angles with respect to the object's coordinate frame.
%We  and then sample the azimuth and elevation angles. 
%We verified that the estimated variance proportions and dimensions were the same when sampling 36 and 120 azimuth (note that the sampling influence the absolute value of the variance). 
%\bryan{The previous sentence does not make sense.} 
Finally, we vary the scene lighting such that the source light varies from left to right and front to back on a uniform grid. % \bryan{Are there other directions?}.
%We sample \bryan{FILL IN} light positions.

For the quantitative experiments we sampled 36 azimuth angles (keeping the elevation fixed at 10 degrees) for rotation, 36 positions for translation, 40 scales, 36 light positions, and 125 colors. 
For the visualizations we sampled 120 azimuth angles, 121 light positions, and 400 positions to make the embeddings easier to interpret. 
We checked that the different sampling did not change our quantitative results, which was expected since our method is not sensitive to the relative number of samples for each factor. %the deformation as soon as the space is sufficiently covered. 

\begin{table}[t]
\center
\caption{
Relative variance of the aspect ratio, 2D position, and residual feature for our synthetic rectangle experiment with AlexNet. 
Notice that the relative variance of the aspect ratio increases with the higher layers while 2D position decreases, which indicates that the features focus more on the shape and less on the 2D location in the image. 
}
{
\begin{tabular}{ | l | c | c | c|}
 \cline{2-4}
\multicolumn{1}{c|}{} &
 2D position & Aspect ratio & $\Delta^L$\\
  \hline\hline%
  AlexNet, pool5 & 49.8 \% & 9.5 \% & 40.8 \% \\
 \hline%\cline{2-4}
   AlexNet, fc6 &  45.1 \% & 22.3 \% & 32.6 \% \\
 \hline%\cline{2-4}
   AlexNet, fc7    & 33.9 \% & 37.0 \% & 29.1 \%  \\ 
\hline
  \end{tabular}
}
\label{tab:rectangle}
\end{table}

\begin{figure*}[t]
\centering
\begin{subfigure}[b]{0.23\linewidth}
\centering
       \includegraphics[width=0.98\linewidth]{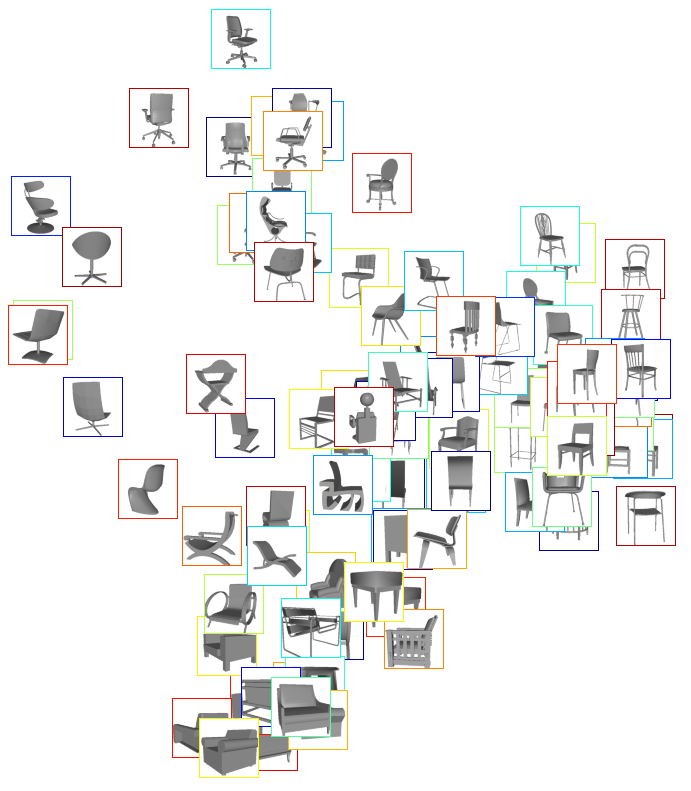}\vspace*{2mm} 
        \caption{Chair, pool5}%, dim. 1 and 2}
\end{subfigure}%
\begin{subfigure}[b]{0.24\linewidth}
\centering
	\hspace{2mm}
       \includegraphics[width=0.98\linewidth]{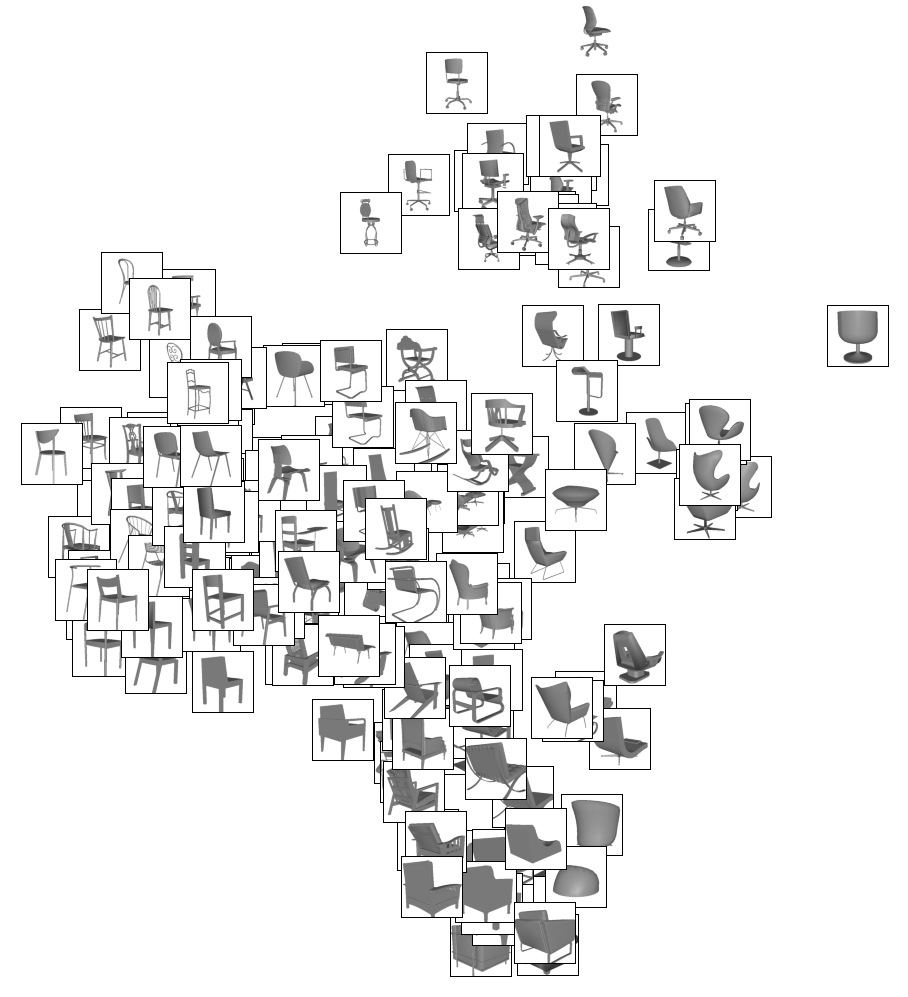}
        \caption{Chair, pool5, style}%, dim. 1 and 2}
\end{subfigure}%
\begin{subfigure}[b]{0.23\linewidth}
\centering
       \includegraphics[width=0.8\linewidth]{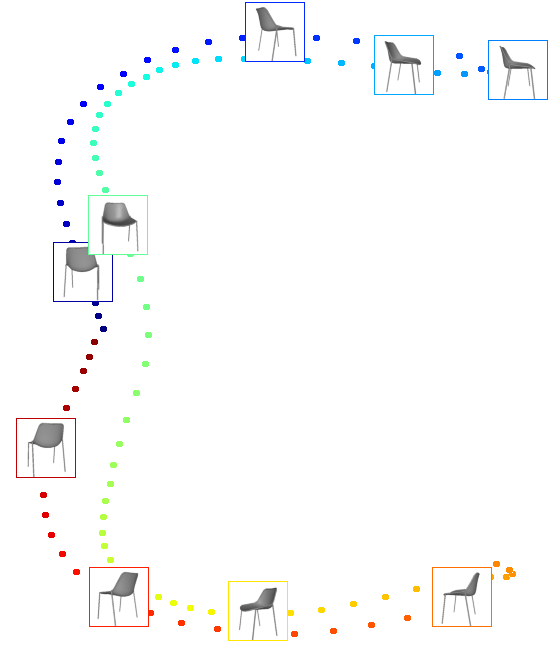}
        \caption{Chair, pool5, rotation}%, dim. 1 and 2}
        \label{fig:chair_rot}
\end{subfigure}%
\begin{subfigure}[b]{0.26\linewidth}
\centering
       \includegraphics[width=0.8\linewidth]{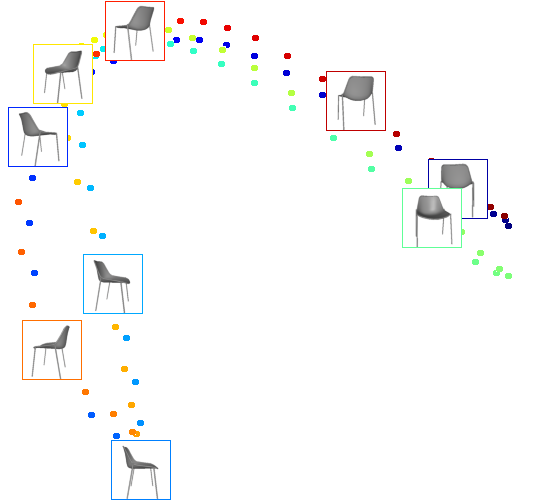}
        \caption{Chair, fc6, rotation}%, dim. 1 and 2}
\end{subfigure}\\%\\
\begin{subfigure}[b]{0.24\linewidth}
\centering
       \includegraphics[width=0.98\linewidth]{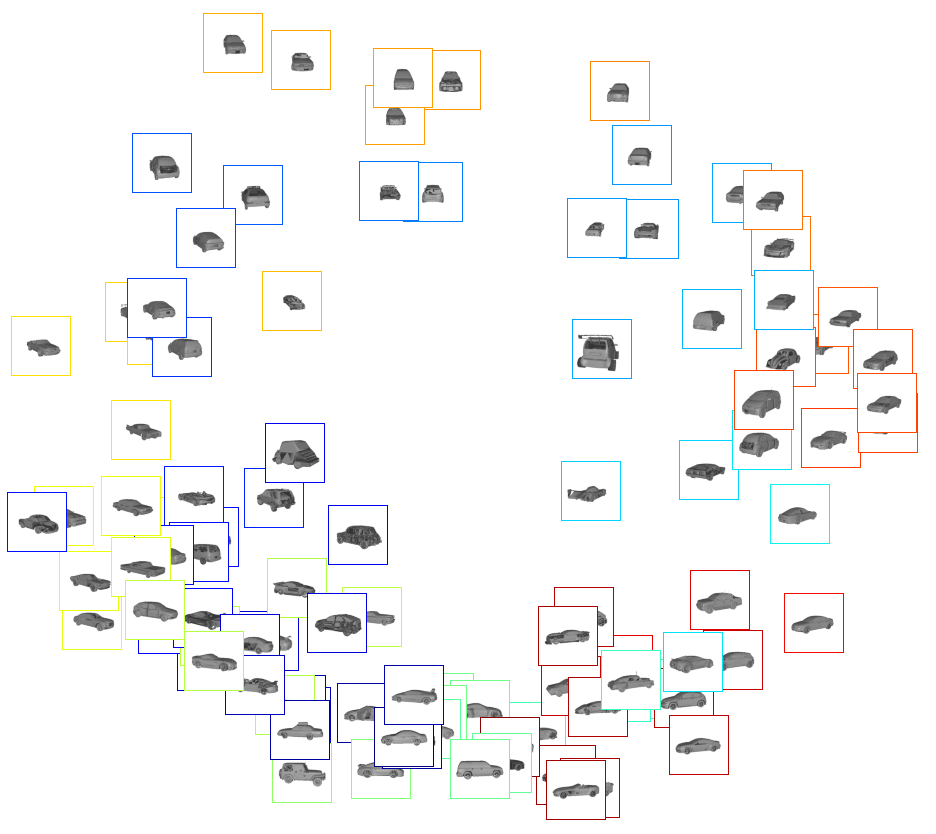}\vspace*{2mm} 
        \caption{Car, pool5}%, dim. 1 and 2}
\end{subfigure}%\\\\ 
\begin{subfigure}[b]{0.20\linewidth}
\centering
	\hspace{2mm}
       \includegraphics[width=0.98\linewidth]{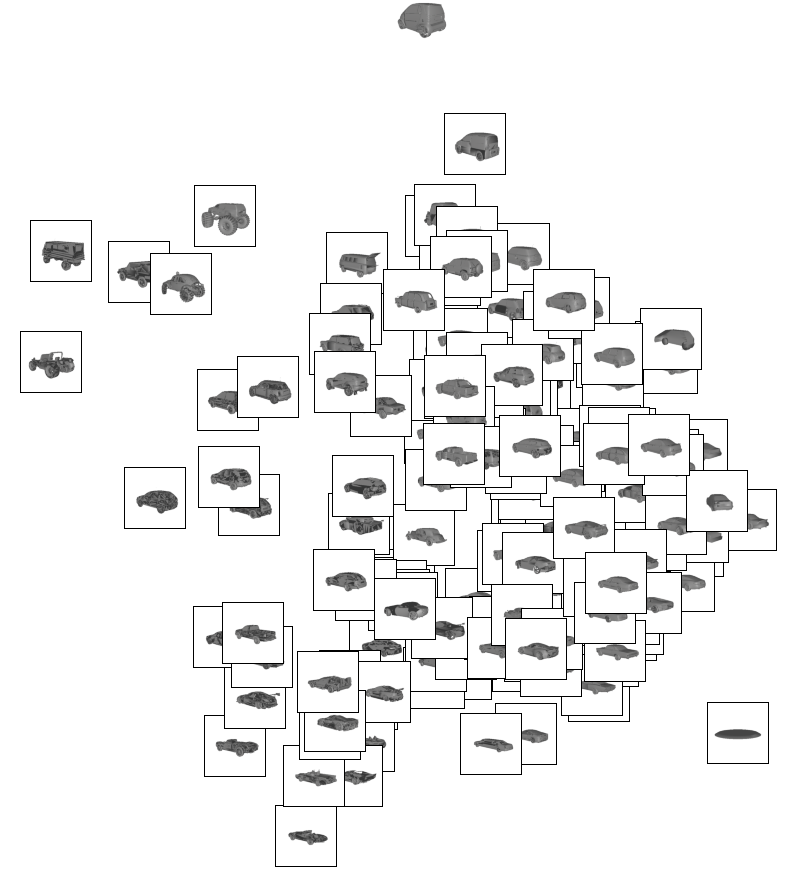}
        \caption{Car, pool5, style}%, dim. 1 and 2}
\end{subfigure} %\\
\begin{subfigure}[b]{0.28\linewidth}
	\hspace{4mm}
       \includegraphics[width=0.8\linewidth]{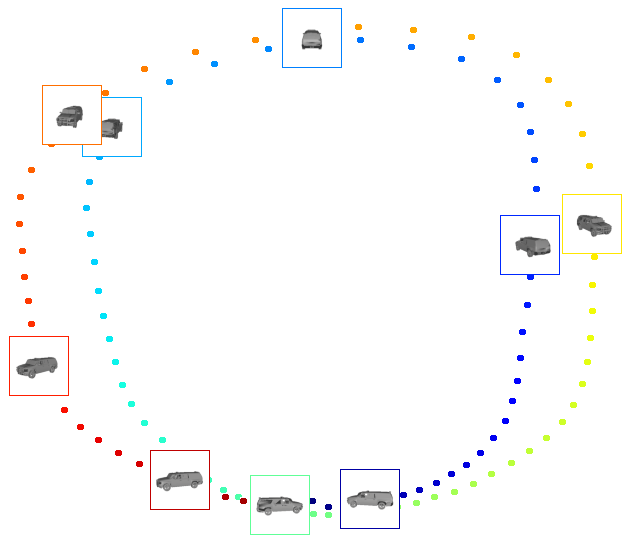}
        \caption{Car, pool5, rotation}%, dim. 1 and 2}
        \label{fig:car_rot}
\end{subfigure}%\\
\begin{subfigure}[b]{0.25\linewidth}
       \includegraphics[width=0.8\linewidth]{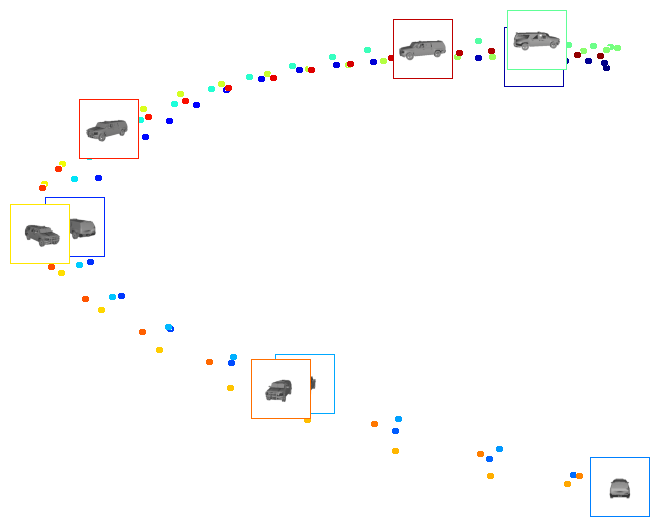}
        \caption{Car, fc6, rotation}%, dim. 1 and 2}
\end{subfigure}%\\
\caption{
{\bf Best viewed in the electronic version.} 
PCA embeddings (dims.\ 1,2) of AlexNet features for ``chairs'' (first row) and ``cars'' (second row). 
Column 1 -- Direct embedding of the rendered images without viewpoint-style separation. 
Columns 2,3 -- Embeddings associated with style (for all rotations) and rotation (for all styles). 
Column 4 -- Rotation embedding for fc6, which is qualitatively different than pool5.  
Colors correspond to orientation and can be interpreted via the example images in columns 3,4. 
Similar results for other categories and PCA dimensions are available in the supplementary material.
} 
\label{fig:all}
\end{figure*}

\section{Results}

\label{sec:results}
In this section we highlight a few results from our experiments on CNN feature analysis.
The supplementary material  reports our detailed quantitative results for all object categories and provides a visualization tool to interactively select and compare embeddings for the first ten PCA components.

We first report results for manually-designed 2D stimuli in section~\ref{sec:2d} and then for rendered views of 3D models from several object categories in section~\ref{sec:3d}. 
We finally show in section~\ref{sec:natural_images} that our results obtained with computer-generated images are related to natural images.%, in particular to perform efficient instance recognition.

\subsection{2D abstract stimuli}
\label{sec:2d}
In this section we apply the deep feature analysis of section~\ref{sec:theorie} on manually-designed 2D abstract stimuli presented to a trained CNN. 
We first perform a PCA analysis on a single factor, color.
Next, we perform two-factor quantitative analyses on the aspect ratio/2D position of a rectangle and on the foreground/background colors of a centered square.
%We first consider color as a single factor, followed by aspect ratio and 2D position of a rectangle as two joint factors.
%We begin by looking at the embedding of manually-designed simple 2D stimuli. 
%The simple 2D stimuli allows simple cases we introduce the use and the interest of the tools presented in section \ref{sec:theorie}. 
%We investigate how uniform color images and images depicting rectangles at different positions and of aspect ratios are represented in a CNN.

\paragraph{Uniform color.} 
%We first look at the embedding of color. 
We perform PCA on a set of images with constant color, as described in section \ref{sec:single_factor}. We sampled 1331 colors uniformly on a grid in RGB color space. The resulting embedding for the fc7 layers of the three CNNs are shown in figure \ref{fig:color}. 
The resulting embeddings for the AlexNet and Places CNNs are surprisingly similar to a hue color wheel, with more variation visible for blues and greens and less for reds and violets. 
%\bryan{The hue/saturation wheel may also appear when PCA is performed on natural images.  We should try to find a reference for this if possible.} 
%\bryan{Is the next sentence true?} 
VGG has a different behavior, with the first dimension similar to saturation. 
The embedding corresponding to the second and third dimensions is similar to that of the other CNNs. %For all CNNs the embedding was similar for the pool5 and fc6 layers. 
%\bryan{Previous sentence not clear.  Hue/saturation wheel?  Or different?} 
%\bryan{Perhaps remove the next sentence since it is redundant with the section intro?} Our supplementary material provides tools to visualize these embeddings for different dimensions and layers. 
%\bryan{The next sentence does not make sense to me.}
The number of PCA dimensions necessary to explain 95\% of the variance is approximately 20 for all three networks and all three layers, which is higher than the three dimensions of the input data.

\paragraph{Position and aspect ratio.} 
We used the methodology of section \ref{sec:multiple_factors} to 
study the features representing a small black rectangle located at different positions with different aspect ratios on a white background. 
The rectangle area was kept constant at $0.26^ 2$ of the image area. 
We consider the position and aspect ratio as two factors of variation and sample 36 positions on a grid and 12 aspect ratios on a log scale. The variance explained by each factor for the different layers of AlexNet is presented in table~\ref{tab:rectangle}. 
For all three networks the relative variance associated to the position decreases, which quantitatively supports the idea that the higher layers have more translation invariance. 
In contrast the relative variance associated to the aspect ratio increases for the higher CNN layers.  
For AlexNet, less than 10\% of the relative variance for pool5 is explained by the aspect ratio alone, while it explains 37\% of the relative variance for fc7. 
Also, the relative variance associated with the residual decreases for the higher CNN layers, which indicates the two factors are more easily linearly separated in the higher layers.
%Note that this was not the case for the bi-color square experiment.

\paragraph{Center and surrounding color.} 
Similar to the experiments presented in the previous paragraph, we considered a square of one color on a background of a different color. 
We chose the size of the central square to be half the image size. 
Quantitative results for the fc7 features of the different networks are presented in table \ref{tab:bicolor}.
Results for the other layers are in the supplementary material.
A first observation is that the features do not separate as well the foreground and background colors in the representation as the aspect ratio and position in the previous experiment. 
%This may be related to a relative interpretation of the colors. 
We also observe that for all the networks the variance associated to the background color is higher than the variance associated to the foreground. 
The difference is more striking for the Places fc7 layer (3.8x versus 2x for AlexNet fc7 and 1.8x for VGG fc7). 
Future work could determine if the background color of an image is especially important for scene classification, while the foreground color is less important.

\begin{table}[t]
\center
\caption{
Relative variance and intrinsic dimensionality of a foreground square of one color on a background color.
Each cell: top -- rel.\ variance; bottom -- intrinsic dim.
%This table report the relative variance (first line) and dimension (second line) of our center color, surrounding color, and residual feature. The dimensions reported corresponds to the PCA dimensions necessary to explain more than 95\% of the variance.
}
{
\begin{tabular}{l | c | c | c|}
\cline{2-4}
& \multirow{1}{*}{\centering Foreground} &\multirow{1}{*}{\centering Background} & \multirow{1}{*}{$\Delta^L$} \\ \hline\hline
\multirow{2}{*}{Places, fc7}& 13.4\%&51.1\%&35.5\%\\
&13&14&216
\\ \hline
 \multirow{2}{*}{AlexNet, fc7}& 19.2\% & 39.9\% & 40.8\% \\
&14&16&315
\\ \hline
\multirow{2}{*}{VGG, fc7}&20.2\%&36.9\%&42.9\%\\
&11&15&216
\\ \hline
\end{tabular}
}
\label{tab:bicolor}
\end{table}

\paragraph{Remarks.}
As the CNNs were not trained on the 2D artificial stimuli presented in this section, we find it somewhat surprising that the embeddings resulting from the above feature analysis is meaningful. 
From our experiments %\mathieu{similar to neuro-psychology experiments? ref?} 
we saw that the CNNs learn a rich representation of colors, identifying in particular variations similar to hue and saturation.
Moreover, the last layers of the network better encode translation invariance, focusing on shape. These results will be confirmed and generalized on more realistic stimuli in the next sections.% \mathieu{speak about analogy with neurological experiments? remove this paragraph?}

\subsection{Object categories}
\label{sec:3d}
%The images studied in the previous section, even if they led to interesting observations, were very different from natural images. 
In this section we want to explore the embedding generated by the networks for image sets and factors related to the tasks for which they are trained, namely object category classification in the case of AlexNet and VGG. 
We also compare against the CNN trained on Places.
We thus  select an object category and, using rendered views of 3D models, we analyze how the CNN features are influenced by the style of the specific instances as well as different transformations and rendering parameters. The parameter sampling for each experiment is described in section \ref{sec:3D_factors}.
%\bryan{The next three sentences is redundant with the intro.} 
%The main difficulty for this study is the access to labeled data. Indeed, while it is possible to have access to a few thousands images for a given category, having hundreds of  views of each instance with detailed annotations is much more difficult. We avoid this difficulty by considering database of 3D models instead of images and rendering views of 3D CAD models with the parameters required for our experiments.
%, focusing mainly on the two categories which have arguably received the most attention in shape analysis: cars \cite{}, and chairs \cite{}. \mathieu{Bryan, despite our discussion, I am not sure how to rewrite this sentence, can you have a try?} . Detailed results for three other categories are presented in the supplementary material and our quantitative results are averages over the 5 categories. 

\paragraph{Model--orientation separation.}
The first variation we study jointly with style is the rotation of the 3D model. % change in viewpoint.
%Here, we 3D rotate the objects around its vertical axis. We fix the elevation angle to 10 degrees and render different azimuth angles. 
The first column of figure \ref{fig:all} visualizes the PCA embedding of the resulting pool5 features. 
This embedding is hard to interpret because it mixes information about viewpoint (important for cars) and instance style (important for chairs). 
To separate this information, we perform the decomposition presented in section \ref{sec:theorie}. 
The decomposition provides us with embedding spaces for style and viewpoint and associates to each model and viewpoint its own descriptor. 
We visualize the embeddings in figure \ref{fig:all}; the second column corresponds to style and the third to viewpoint. 
Note that the different geometries of the two categories lead to different embeddings of rotation in pool5.
While a left-facing car typically looks similar to a right-facing car and is close in the feature space (figure \ref{fig:car_rot}), a right-facing chair is usually different from left-facing chairs and is far in the embedding (figure \ref{fig:chair_rot}).  
The last column shows the viewpoint embedding for fc6.
The comparison of the last two columns indicates that much viewpoint information is lost between pool5 and fc6 and that fc6 is largely left-right flip invariant. %\mathieu{probably needs some discussion}
A potential interesting future direction could be to interpret the viewpoint embeddings relative to classic work on mental rotation~\cite{Shepard71}.

\begin{figure}[t]
\centering
\begin{subfigure}[b]{0.45\linewidth}
       \includegraphics[width=0.8\linewidth]{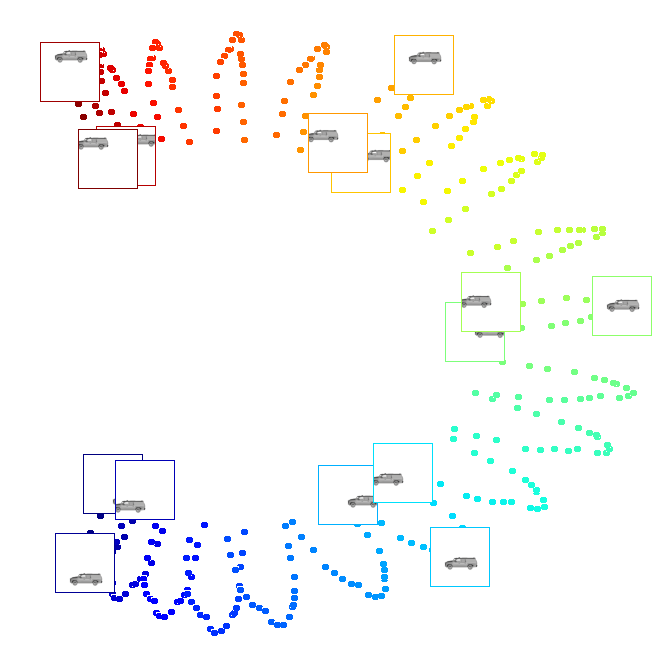}
        \caption{Car, pool5}%, dim. 1 and 2}
\end{subfigure}%
\begin{subfigure}[b]{0.45\linewidth}
       \includegraphics[width=0.8\linewidth]{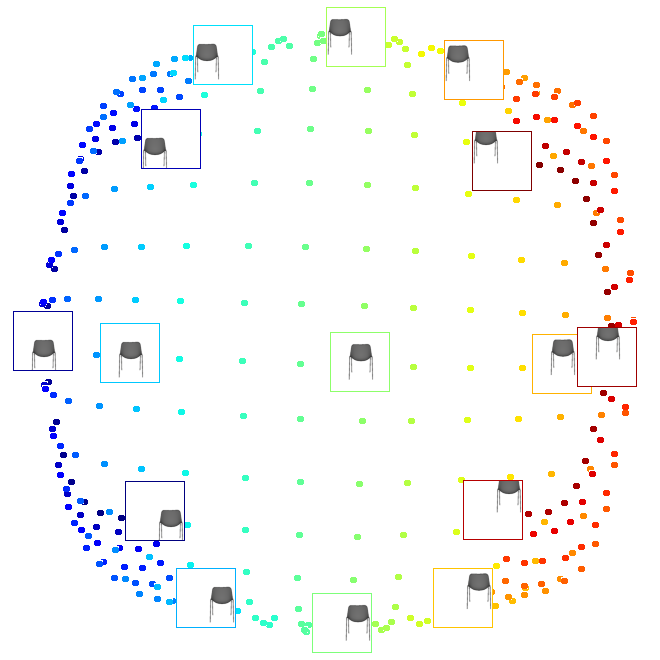}
        \caption{Chair, pool5}%, dim. 1 and 2}
\end{subfigure}\\
\begin{subfigure}[b]{0.45\linewidth}
       \includegraphics[width=0.8\linewidth]{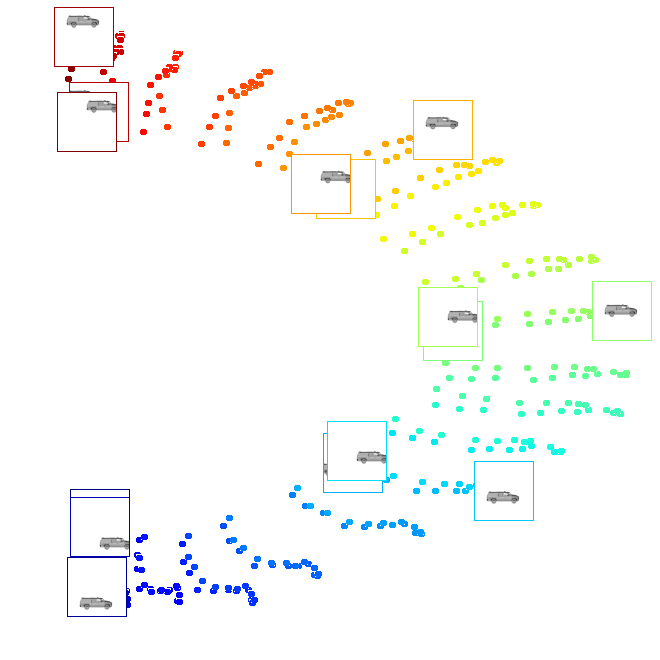}
        \caption{Car, fc6}%, dim. 1 and 2}
\end{subfigure}%
\begin{subfigure}[b]{0.45\linewidth}
       \includegraphics[width=0.8\linewidth]{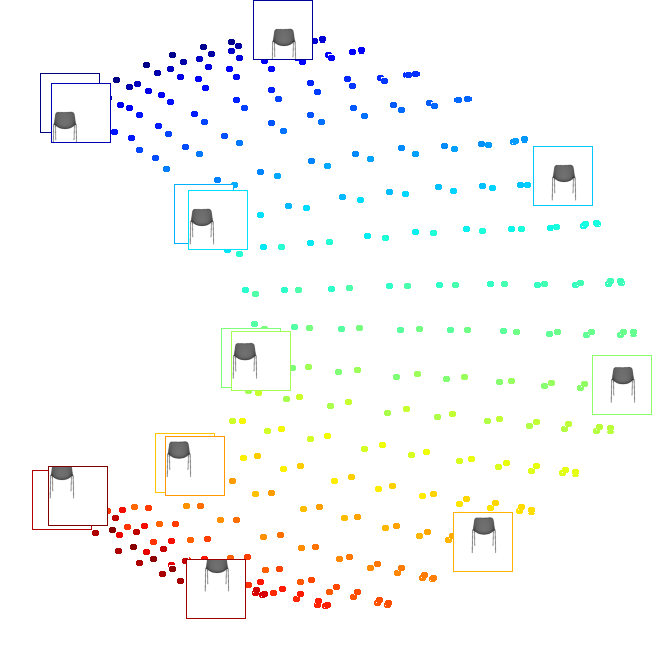}
        \caption{Chair, fc6}%, dim. 1 and 2}
\end{subfigure}\\
\begin{subfigure}[b]{0.45\linewidth}
       \includegraphics[width=0.8\linewidth]{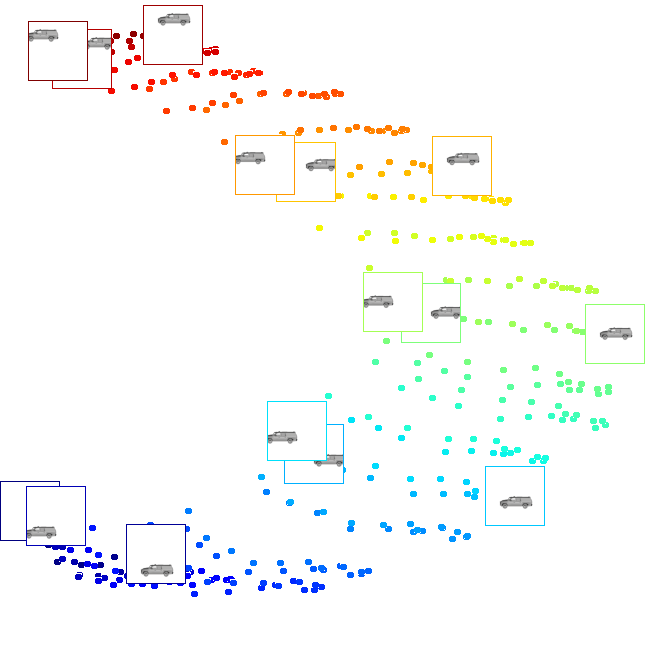}
        \caption{Car, fc7}%, dim. 1 and 2}
\end{subfigure}%
\begin{subfigure}[b]{0.45\linewidth}
       \includegraphics[width=0.8\linewidth]{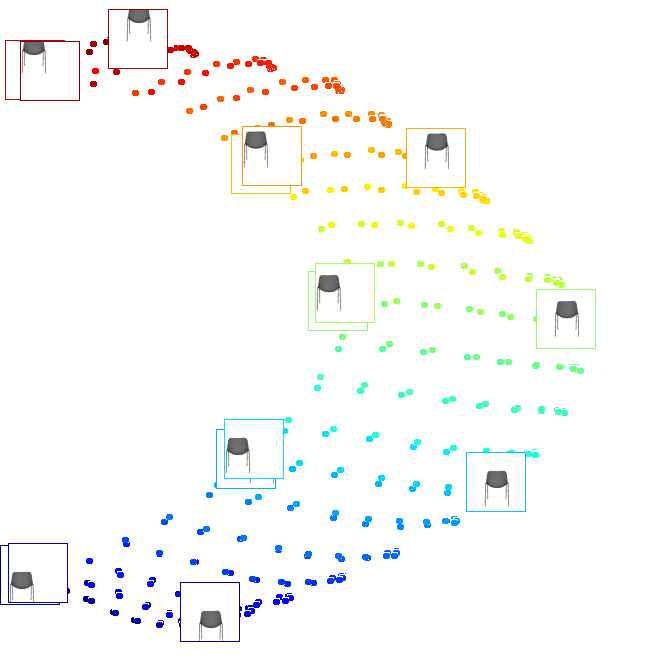}
        \caption{Chair, fc7}%, dim. 1 and 2}
\end{subfigure}%
\caption{
PCA embeddings for 2D position on AlexNet.
%\bryan{We need to explain why (a) and (b) look different.}
} 
\label{fig:translation}
\end{figure}

\begin{figure}[t]
\centering
\begin{subfigure}[b]{0.49\linewidth}
       \includegraphics[width=0.8\linewidth]{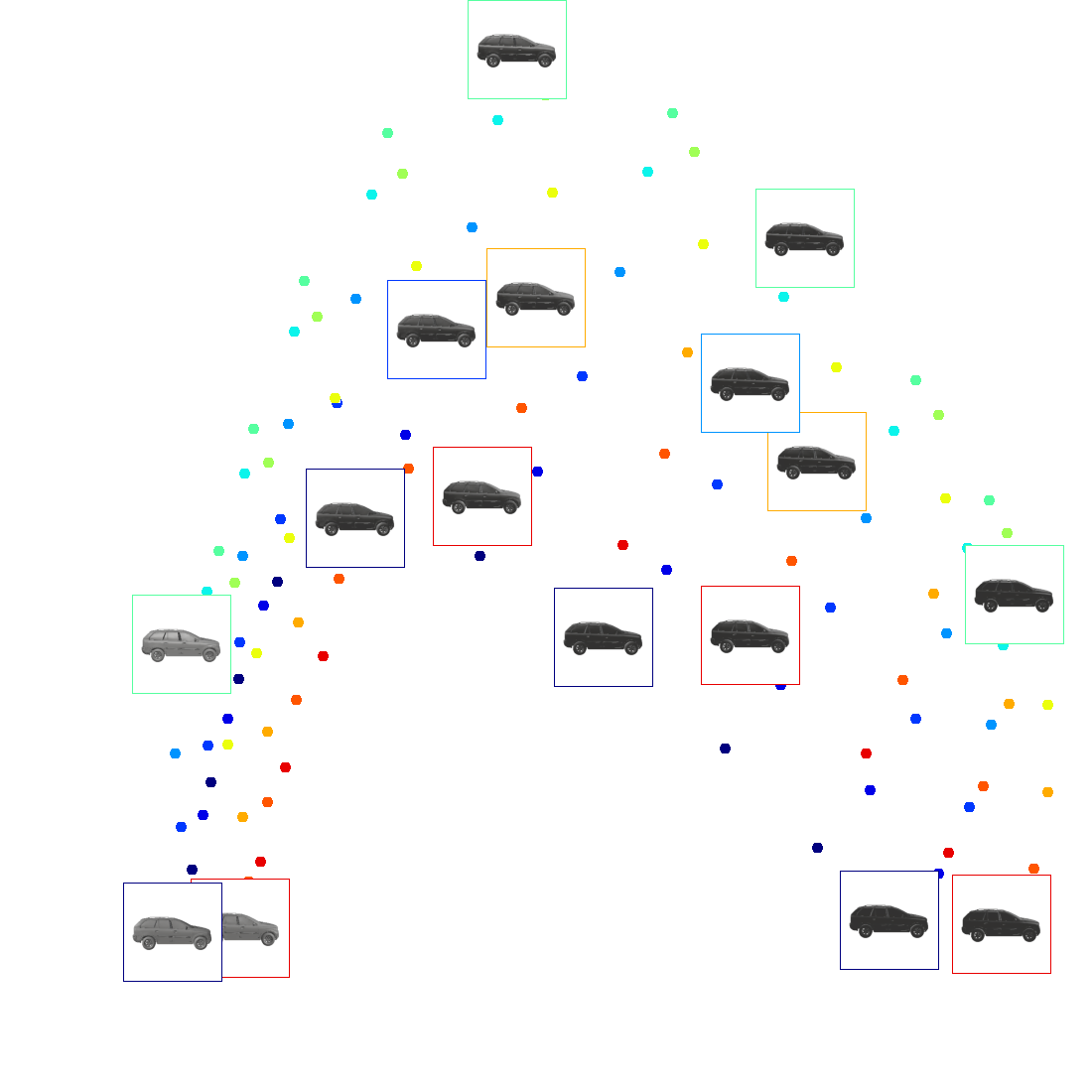}
        \caption{Lighting}%, dim. 1 and 2}
        \label{fig:lightening}
\end{subfigure}%\\
\begin{subfigure}[b]{0.49\linewidth}
       \includegraphics[width=0.8\linewidth]{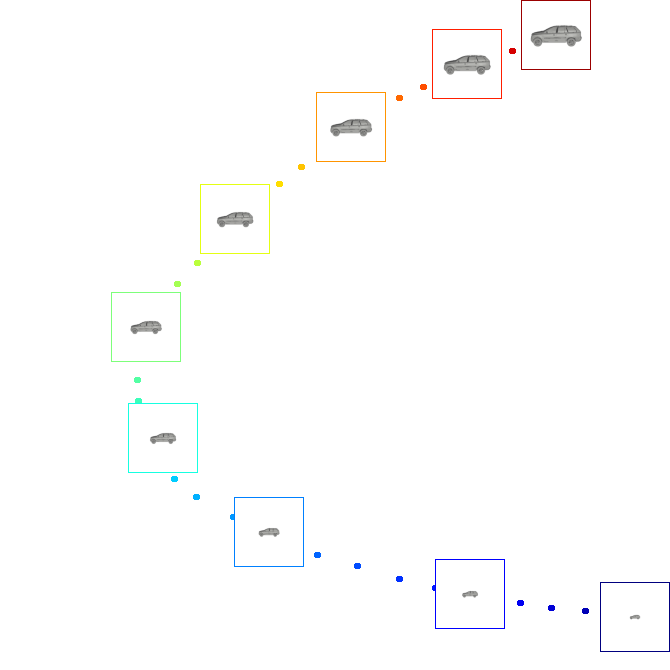}
        \caption{Scale}%, dim. 1 and 2}
\end{subfigure}\\%
\begin{subfigure}[b]{0.49\linewidth}
       \includegraphics[width=0.8\linewidth]{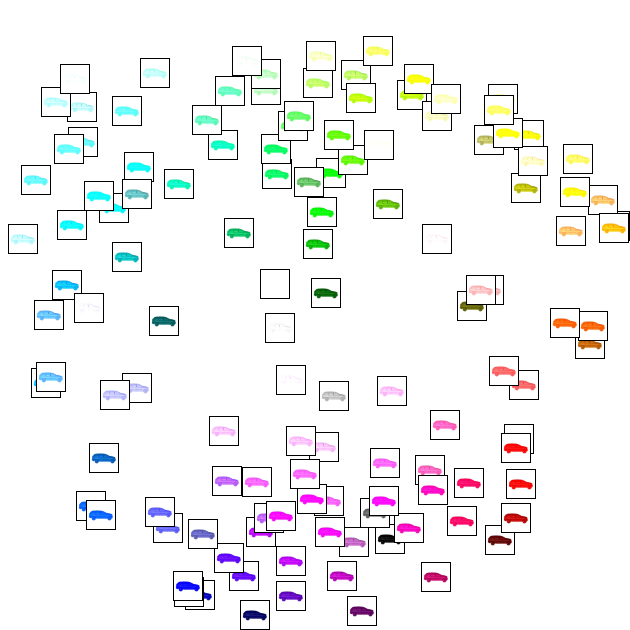}
        \caption{Object color}%, dim. 1 and 2}
\end{subfigure}%\\
\begin{subfigure}[b]{0.49\linewidth}
       \includegraphics[width=0.8\linewidth]{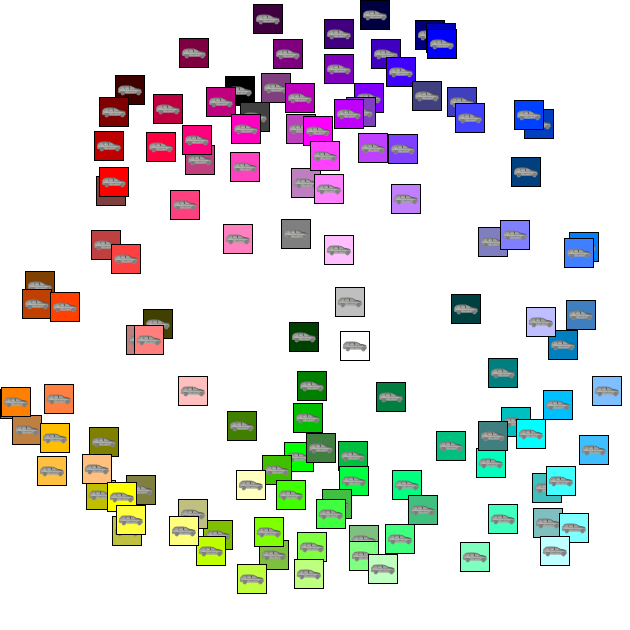}
        \caption{Background color}%, dim. 1 and 2}
\end{subfigure}%
\caption{PCA embeddings for different factors using AlexNet pool5 features on ``car'' images. 
Colors in (a) correspond to location of the light source (green -- center).
} 
\label{fig:other}
\end{figure}

\paragraph{Translation, scale, lighting, color.} 
We repeated the same experiment for the following factors: 2D translation, scale, light direction, background color, and object color. 
For simplicity and computational efficiency, we considered in all experiments a frontal view of all the instances of the objects.
The framework allows the same analysis using the object orientation as an additional factor. 
The embeddings associated with AlexNet features for translation of cars and chairs are shown in figure~\ref{fig:translation}. 
Note that similar to rotations, the embedding corresponding to cars and chairs are different, and that the first two components of the fc6 features indicate a left-right flip-invariant representation. 
The embeddings for the pool5 layer of the car category for the other factors are shown figure \ref{fig:other}. 
%Note that all these embeddings are easy to interpret, showing both the pertinence of our category-level analysis and the power of the CNN embeddings.
%\bryan{The previous sentence is weak.  Is there anything else to say?}

%\paragraph{Embeddings associated to different rendering parameters:}\mathieu{show figures and images + does color embedding change depending on the category?}

\paragraph{Quantitative analysis: viewpoint.}%\mathieu{needs a table}
We analyze the relative variance explained by the 3D rotation, translation, and scale experiments. 
While the variance was different for each factor and category, the variation across the layers and networks was consistent in all cases. 
For this reason we report in table \ref{table:geometry} an average of the variance across all five categories and all three factors.
We refer the reader to the supplementary material for detailed results. %Similar to the rotation, we study the importance of other rendering parameters: the 2D position of the object, its scale, its illumination, the background and object color. The scale, 2D translation and background color are associated to a part of the variance similar to the one of the azimuth rotation around 10\%, while the lightening is associated to a much smaller variation, around 1\%. Because of the very small varince associated to the lightening conditions, we neglect it in our next experiments. The analysis of the dimensionality of the features reveal more interesting informations: the features associated to 1D translation and scale have a dimension around 2, the 2D translation around 10 and the background color around 17. This relative richness of the color embedding might explain a part of the superiority of CNN over hand designed features, which typically had a very poor representation of colors, which the are actually very related to the scene content \cite{extended_Hoogles}
The analysis of table \ref{table:geometry} reveals several observations. First, the proportion of the variance of deeper layers corresponding to viewpoint information is less important, while the proportion corresponding to style is more important. 
This corresponds to the intuition that higher layers are more invariant to viewpoint.
We also note that the residual feature $\Delta^L$ is less important in higher layers, indicating style and viewpoint are more easily separable in those layers. 
These observations are consistent with our results of section~\ref{sec:2d}.
Second, the part of the variance associated with style is more important in the fc7 layer for VGG than in AlexNet and Places.
Also, the part associated with the viewpoint and residual is smaller. 
Note that this does not hold in pool5, where the residual is important for the VGG network. 
This effect may be related to the difference in the real and intrinsic dimension of the features. 
The intrinsic dimension of the style component of VGG pool5 features is larger and decreases from pool5 to fc7. 
On the contrary, the intrinsic dimensionality of AlexNet has smaller variation across layers. 
Finally, we note that the intrinsic dimensionality of the fc7 style feature of Places is smaller than the other networks. 
This may indicate that it is less rich, and may be related to the fact that identifying the style of an object is less crucial for scene classification.
We believe it would be an interesting direction for future work to study how the improved performance of VGG for object classification is related to the observed reduced sensitivity to viewpoint.

\begin{table}
\center
\caption{
Relative variance and intrinsic dimensionality averaged over experiments for different object categories and viewpoints (3D orientation, translation, and scale).
Each cell: top -- rel.\ variance; bottom -- intrinsic dim.
We do not report the intrinsic dim.\ of $\Delta^L$ since it is typically larger than 1K across the experiments and expensive to compute.
}
{
\begin{tabular}{c |  l | c | c | c|}
 \cline{3-5}
\multicolumn{2}{c|}{}& pool5 & fc6 & fc7\\
  \hline\hline
& Places & 26.8 \% &  21.4  \% & 17.8  \%\\  
  &&  8.5 &  7.0  &   5.9 \\ \cline{2-5}
 Viewpoint   & AlexNet & 26.4  \% &  19.4   \% & 15.6   \%\\ 
 &&   8.3   &    7.2 &  6.0 \\   \cline{2-5}
 & VGG & 21.2   \% &  16.4   \% & 12.3  \%\\  
 &&   10.0  &    7.7   &   6.2 \\   \hline \hline
& Places &  26.8  \% &   39.1  \% & 49.4  \%\\ 
 &&      136.3 &   105.5  &  54.6  \\  \cline{2-5}
 Style& AlexNet &  28.2 \% &  40.3   \% & 49.4   \%\\  
 &&    121.1   &   125.5   &   96.7  \\  \cline{2-5}
 & VGG &  26.4 \% &   44.3    \% &  56.2 \%\\   
 &&    181.9  &    136.3  & 94.2 \\ \hline\hline
 & Places &  46.8    \% & 39.5   \% &   32.9   \%\\  
% &&    -   &   -   &   -   \\
 \cline{2-5}
 $\Delta^L$ & AlexNet &  45.0     \% &   40.3    \% &    35.0  \% \\
% &&   -    &   -    &    -  \\  
 \cline{2-5}
 & VGG &   52.4  \% &  39.3    \% &     31.5 \%\\ 
% &&   -    &  -     &  -  \\    
\hline \hline
  \end{tabular}
}
\label{table:geometry}
\end{table}

\paragraph{Quantitative analysis: color.} 
We report in table \ref{table:color} the average across categories of our quantitative study for object and background color. 
The results are different from those of viewpoint. 
First, we observe that a larger part of the variance of the features of the Places network is explained by the color in all layers. 
This may be related to the fact that color is a stronger indicator of the scene type than it is of an object category. 
Second, while the part of the variance explained by foreground and background color is similar in the fc7 feature of the Places network, it is much larger for the foreground object than for the background object in AlexNet and VGG. Once again, one can hypothesize that it is related to the fact that the color of an object is more informative than the color of its background for object classification. 
Finally, we note that similarly to our previous experiments, the difference between networks is present in pool5 and increases in the higher layers, indicating that the features become more tuned to the target task in the higher layers of the networks.

\begin{table}
\center
\caption{
Average relative variance over five classes for color/style separation. 
}
{
\begin{tabular}{c |  l | c | c | c|}
 \cline{3-5}
\multicolumn{2}{c|}{Foreground/Style}& pool5 & fc6 & fc7\\
  \hline\hline
 & Places &  23.4 \% &  29.4    \% &  34.9 \%\\  
 FG color
 & AlexNet &    23.2 \% &    24.0  \% &  24.0 \%\\ 
 & VGG & 15.0   \% &  22.6   \% & 25.0  \%\\  
  \hline
& Places &   48.9  \% &   41.3     \% & 40.3    \%\\ 
Style
 & AlexNet &   56.6 \% &     52.1 \% &   52.5 \%\\  
 & VGG &  59.0  \% &   51.4    \% &  51.3 \%\\   
\hline
 & Places &      27.7  \% &    29.3  \% &    24.8     \%\\  
$\Delta^L$
 & AlexNet &        20.3 \% &   24.0      \% &     23.5    \% \\
 & VGG &   26.0  \% &  25.9    \% &    23.6 \%\\ 
\hline
\multicolumn{5}{c}{}\\ 
 \cline{3-5}
\multicolumn{2}{c|}{Background/Style} & pool5 & fc6 & fc7\\
  \hline\hline
 & Places &  24.3  \% & 29.6    \% &  35.1 \%\\  
 BG color  
 & AlexNet &    17.3  \% &   16.2  \% &    14.4 \%\\ 
 & VGG & 9.1   \% &  13.8   \% & 14.3  \%\\  
 \hline
& Places &   51.5  \% &    40.7    \% &   40.9 \%\\ 
Style
&  AlexNet &    63.7  \% &    59.4  \% &    61.8 \%\\  
 & VGG &  71.4  \% &   64.3    \% &  65.3 \%\\   
\hline
 & Places &    24.2   \% &    29.7 \% &      24.0   \%\\  
$\Delta^L$
 & AlexNet &       19.0 \% &       24.4  \% &     23.9    \% \\
& VGG &   19.5  \% &  22.0    \% &    20.4 \%\\ 
\hline
  \end{tabular}
}
\label{table:color}
\end{table}

\subsection{Natural images}
\label{sec:natural_images}

\begin{figure}[t]
\centering
\begin{subfigure}[b]{0.49\linewidth}
       \includegraphics[width=0.8\linewidth]{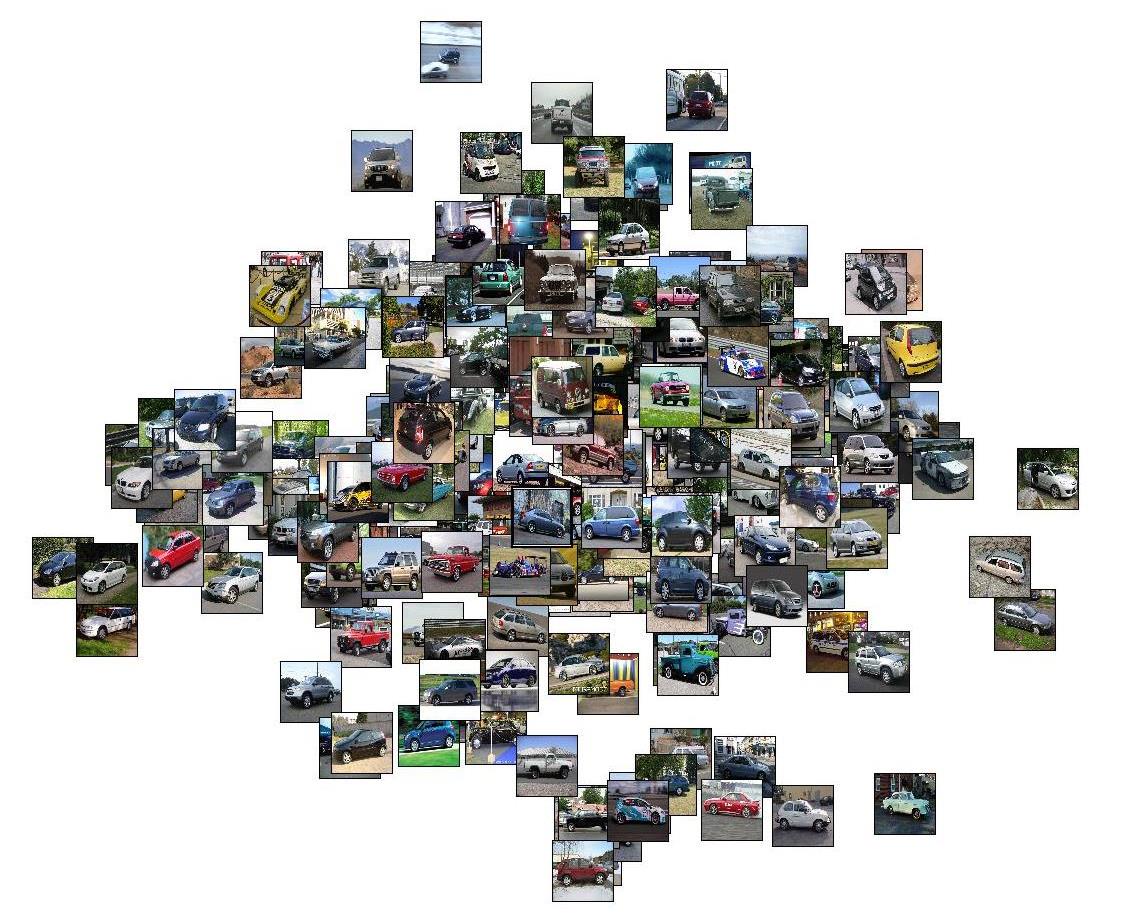}
        \caption{Car orientation}%, dim. 1 and 2}
\end{subfigure}%\\
\begin{subfigure}[b]{0.49\linewidth}
       \includegraphics[width=0.8\linewidth]{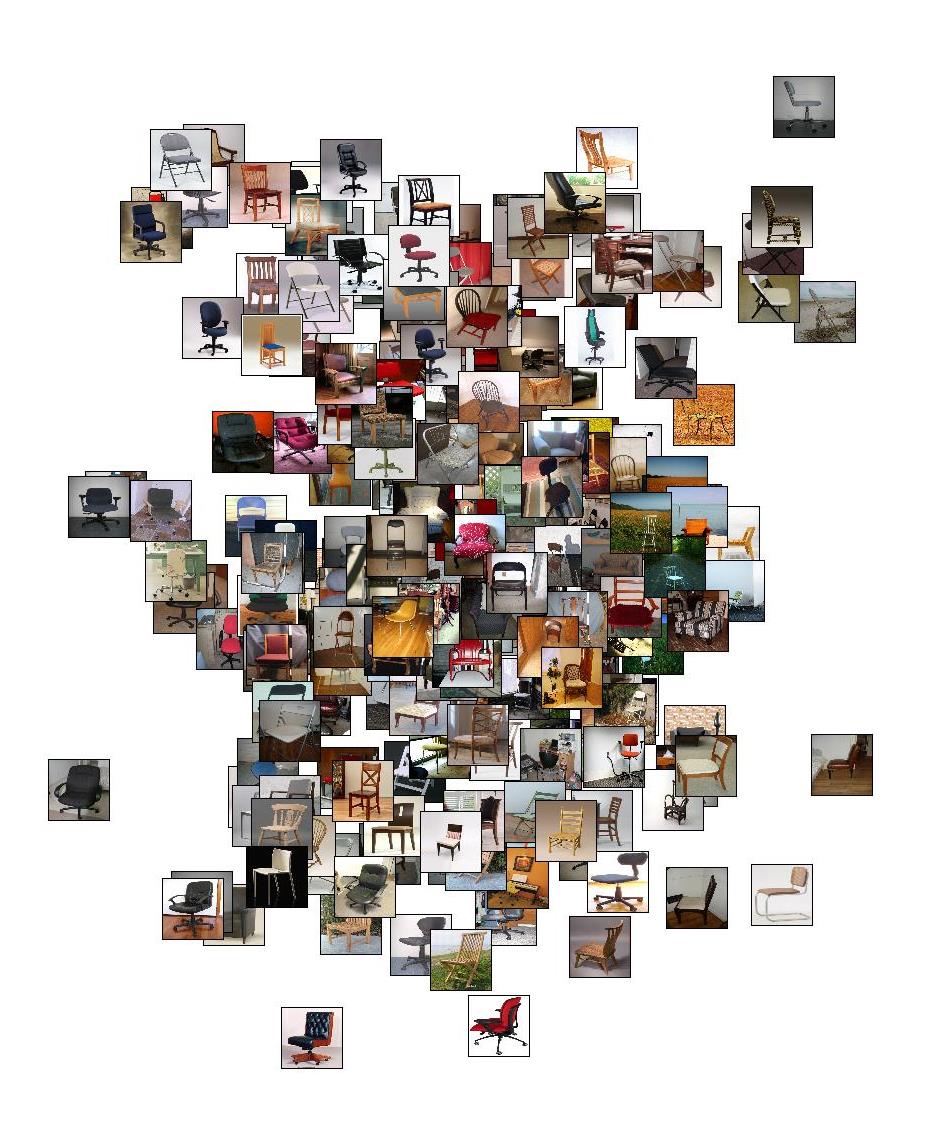}
        \caption{Chair orientation}%, dim. 1 and 2}
\end{subfigure}\\%
\begin{subfigure}[b]{0.49\linewidth}
       \includegraphics[width=0.8\linewidth]{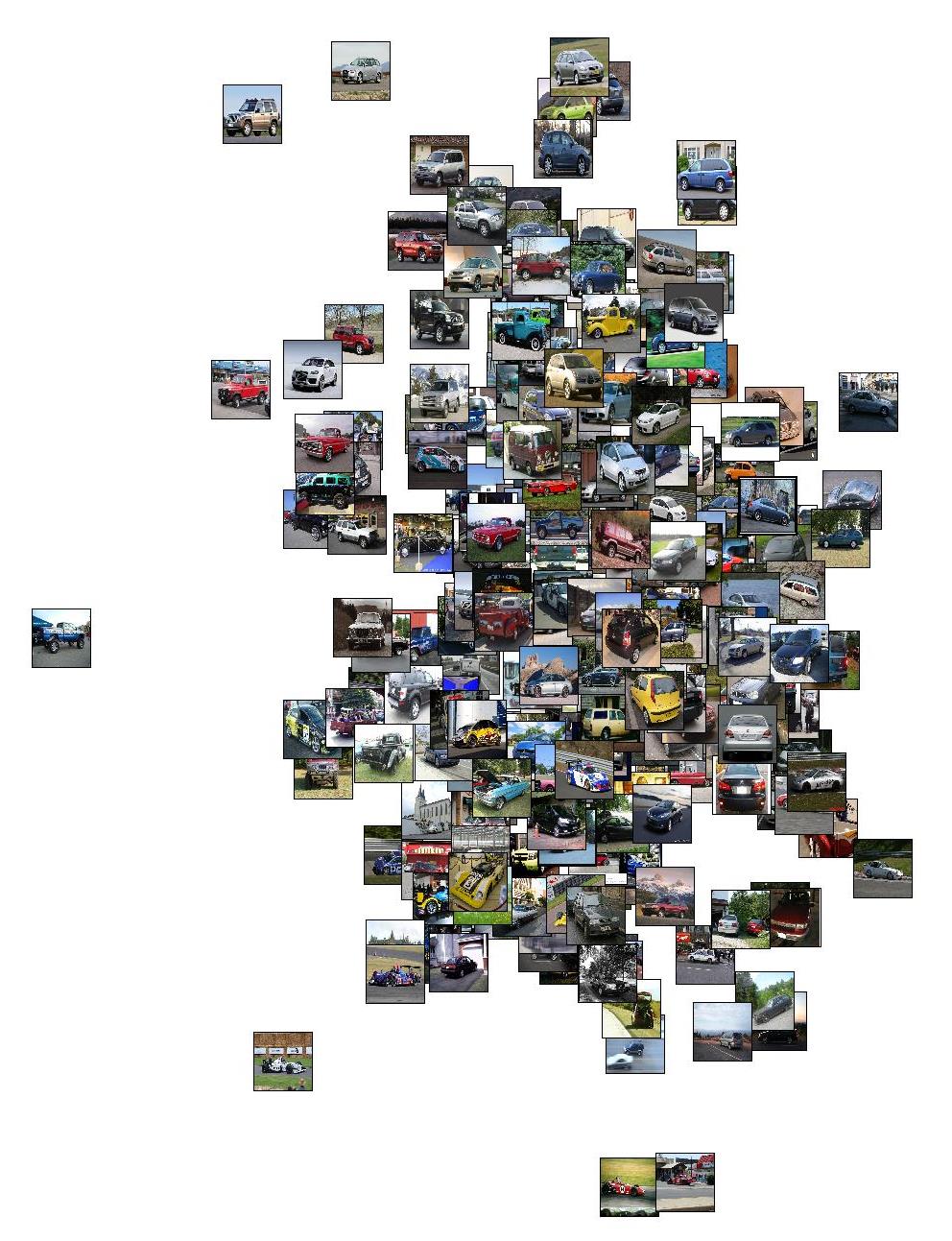}
        \caption{Car Style}%, dim. 1 and 2}
\end{subfigure}%\\
\begin{subfigure}[b]{0.49\linewidth}
       \includegraphics[width=0.8\linewidth]{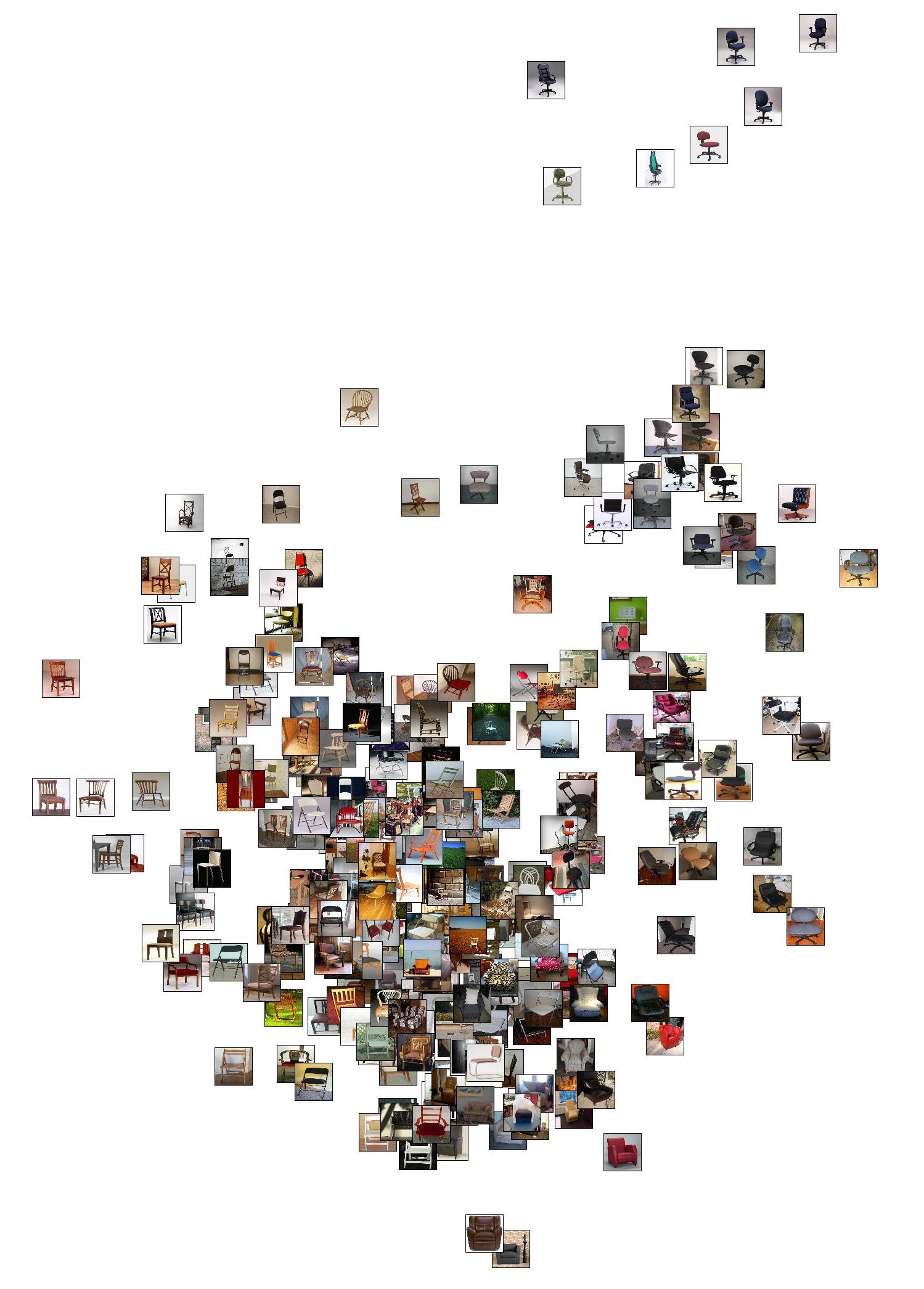}
        \caption{Chair style}%, dim. 1 and 2}
\end{subfigure}\\
\begin{subfigure}[b]{\linewidth}%, dim. 1 and 2}
        \includegraphics[width=0.08\linewidth]{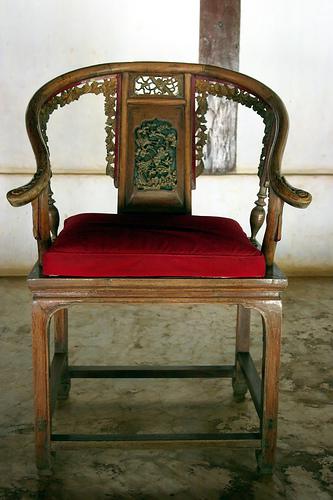}
       \includegraphics[width=0.12\linewidth]{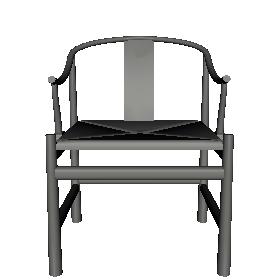}
       \includegraphics[width=0.08\linewidth]{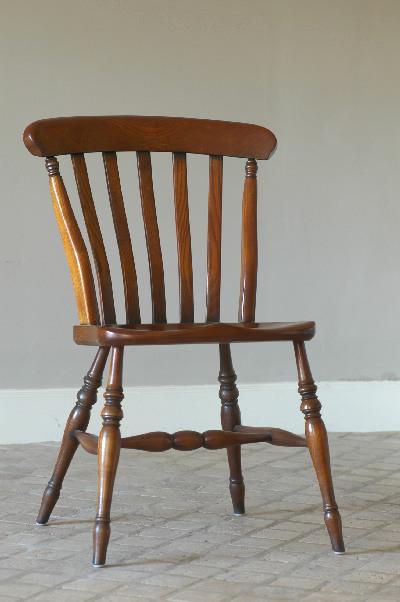}
       \includegraphics[width=0.11\linewidth]{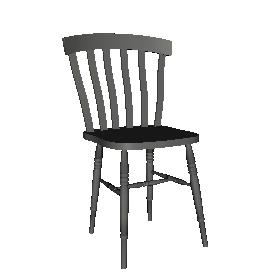}
       \includegraphics[width=0.115\linewidth]{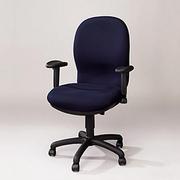}
       \includegraphics[width=0.115\linewidth]{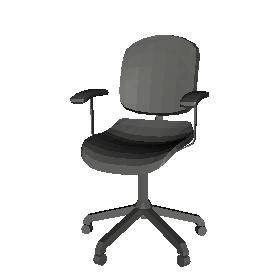}
       \includegraphics[width=0.135\linewidth]{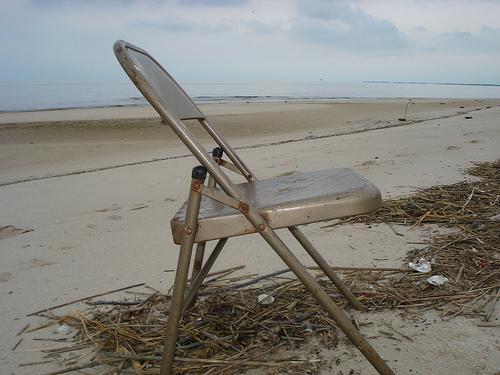}
       \includegraphics[width=0.105\linewidth]{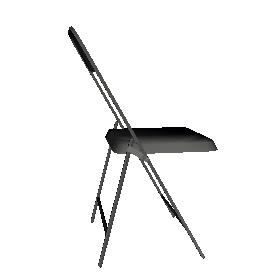}
        \caption{2D-3D retrieval examples}
\label{fig:nn}
\end{subfigure}%
\caption{
PCA embeddings over AlexNet pool5 features for cars and chairs with orientation and style separated.
} 
\label{fig:real_embedding}
\end{figure}
\paragraph{Embedding.} We used ImageNet \cite{ILSVRCarxiv14} images to study the embeddings of natural images. 
Since we have no control over the image content, we cannot perform a detailed analysis of the different factors similar to the previous sections.
Our only choice is to consider the images altogether. 
The direct embedding of natural images is possible but hard to interpret. 
We can however project the images in the spaces discovered in section \ref{sec:3d}. 
The resulting embeddings for style and viewpoint are shown in figure \ref{fig:real_embedding} and are similar to the embeddings obtained with the CAD models.

\paragraph{2D-3D instance recognition.} 
The observed similarity of the embeddings for natural and rendered images motivates an application to retrieve a 3D model of an object present in an image without explicitly performing alignment or detection. 
The approach is different from approaches in the 2D-3D matching community that find explicit correspondences between the models and the image. 
We tested this idea using the chair category and computing similarity as dot product between AlexNet features. 
We rendered 36 azimuth and 4 elevation angles to span typical viewpoints depicted in the natural images.
To improve efficiency, we reduced the dimension of our features to 1000 via PCA. 
This allows us to perform nearest-neighbor retrieval between 1000 natural images and the 144 rendered views of 1261 3D models in approximately 22 seconds total using a Python implementation with pre-computed features.
%Without feature computation time, the rendered models features can be pre-computed and the image features computation which took 21 seconds in our simple  implementation.  
We visualize results in figure \ref{fig:nn} and in the supplementary material. 
We evaluated the viewpoint accuracy using the annotations of \cite{Xiang14} and found that the orientation error was below 20 degrees for 60\% of the images using pool5 features, 39\% using fc6, and 26\% using fc7. 
This is consistent with our earlier finding that orientation is not as well represented in the higher layers. 
We conducted a user study on Mechanical Turk to evaluate the quality of the style matching for the images where the orientation was correctly estimated with the pool5 features. 
The workers were presented with a pair of images and asked to judge if the style of the chairs and their orientation were similar or different, similar to \cite{Aubry14b}. 
Each pair was evaluated 5 times.
There was agreement in 75\% of the cases that the match was fair and in 18\% that it was exact. 
While the above results could probably be beaten by state-of-the-art 2D-3D matching techniques or simply by adding position and scale to our database, they show that our analysis of rendered 3D models is pertinent for understanding the CNN representation of natural images. %The results are presented table.\mathieu{thinking about it I think it is important to have at least some of it because in addition of being cool it proves that all the things we do in the previous paragraph are not stupid and decorrelated from the natural images}

\paragraph{Real images with pose variation.} 
We applied the methodology of section \ref{sec:3d} to natural images in the ETH-80 dataset \cite{leibe2003}. 
The dataset spans 8 object categories with 10 instances each captured under 41 viewpoints.
As the dataset is significantly smaller than the ModelNet database, this limits considerably the number of variations we can explore. 
Results using AlexNet features are shown in table \ref{tab:eth}.
The high-level conclusions are the same as those of section \ref{sec:3d}, but the differences are less obvious. 
The variance is explained more by the style and less by the viewpoint and residual as one progresses toward the higher layers in the network. 
Please see detailed results for each category in the supplementary material.

%In an alternative approach to the use of synthetic images and the link made between real and synthetic images in the previous paragraphs by comparing their features, we applied the same methodology as in section \ref{sec:3d} to real images. As stated in the introduction, this limits considerably the number of variations one can explore, and in particular only a few instances are typically present for each category. We present results on the ETH-80 dataset \cite{leibe2003} (8 categories with 10 instances each seen under 41 different viewpoint angles) with AlexNet in table \ref{tab:eth}.
%The high level conclusions are the same as those of the section \ref{sec:3d} of the paper, but the differences are less obvious. The variance is explained more and more by the category and less and less by the viewpoint as one progresses in the network. Please see the detailed results for each category in the supplementary material.

\begin{table}[t]
\center
\caption{
Average relative variance over the 8 categories of the ETH-80 dataset~\cite{leibe2003}. 
%While the results may vary for the different categories, they present in general the same effects as the average presented in this table.
%The importance of the residual and viewpoint decreases with the layer, while the importance of the style increases.
}
{
\begin{tabular}{ | l | c | c | c|}
\cline{2-4}
\multicolumn{1}{c|}{} &
Rotation & Style & $\Delta^L$\\
\hline\hline%
AlexNet, pool5 & 35.4 \% &21.6 \% &43.0 \% \\
\hline%\cline{2-4}
AlexNet, fc6 & 30.2 \% &27.7 \% & 42.0 \% \\
\hline%\cline{2-4}
AlexNet, fc7 & 29.5 \% & 30.5 \% &40.0 \% \\ 
\hline
\end{tabular}
}
\label{tab:eth}

\end{table}

\section{Conclusion}
We have introduced a method to qualitatively and quantitatively analyze deep features by varying the network stimuli according to factors of interest. 
Utilizing large collections of 3D models, we applied this method to compare the relative  importance of different factors and have highlighted the difference in sensitivity between the networks and layers to object style, viewpoint and color. 
We believe our analysis gives new intuitions and opens new perspectives for the design and use of convolutional neural networks. 
%We plan to release the source code and data used in the experiments.

\paragraph{Acknowledgments.}
Mathieu Aubry was partly supported by ANR project Semapolis ANR-13-CORD-0003, Intel, a gift from Adobe, and hardware donation from Nvidia. %We acknowledge brunch for inspiration and sustenance.

{\small
\bibliographystyle{ieee}
\bibliography{shortstrings,refs}

\newcommand{\noopsort}[1]{} \newcommand{\printfirst}[2]{#1}
  \newcommand{\singleletter}[1]{#1} \newcommand{\switchargs}[2]{#2#1}
\begin{thebibliography}{10}\itemsep=-1pt

\bibitem{Agrawal14}
P.~Agrawal, R.~Girshick, and J.~Malik.
\newblock Analyzing the performance of multilayer neural networks for object
  recognition.
\newblock In {\em ECCV}, 2014.

\bibitem{Aubry14b}
M.~Aubry, D.~Maturana, A.~A. Efros, B.~C. Russell, and J.~Sivic.
\newblock Seeing {3D} chairs: Exemplar part-based {2D-3D} alignment using a
  large dataset of {CAD} models.
\newblock In {\em CVPR}, 2014.

\bibitem{Aubry14}
M.~Aubry, B.~C. Russell, and J.~Sivic.
\newblock Painting-to-{3D} model alignment via discriminative visual elements.
\newblock {\em ACM Transactions on Graphics}, 33(2), 2014.

\bibitem{Barrow78}
H.~Barrow and J.~Tenenbaum.
\newblock Recovering intrinsic scene characteristics from images.
\newblock In A.~Hanson and E.~Riseman, editors, {\em Computer Vision Systems},
  pages 3--26. Academic Press, N.Y., 1978.

\bibitem{Berkes06}
P.~Berkes and L.~Wiskott.
\newblock On the analysis and interpretation of inhomogeneous quadratic forms
  as receptive fields.
\newblock {\em Neural Computation}, 2006.

\bibitem{Biederman95}
I.~Biederman.
\newblock {\em Visual object recognition}, volume~2.
\newblock MIT press Cambridge, 1995.

\bibitem{Bruna13}
J.~Bruna and S.~Mallat.
\newblock Invariant scattering convolution networks.
\newblock {\em IEEE PAMI}, 35(8):1872--1886, 2013.

\bibitem{Chatfield14}
K.~Chatfield, K.~Simonyan, A.~Vedaldi, and A.~Zisserman.
\newblock Return of the devil in the details: Delving deep into convolutional
  nets.
\newblock In {\em Proc. BMVC.}, 2014.

\bibitem{Cheung15}
B.~Cheung, J.~Livezey, A.~Bansal, and B.~Olshausen.
\newblock Discovering hidden factors of variation in deep networks.
\newblock In {\em ICLR workshop}, 2015.

\bibitem{Dosovitskiy15}
A.~Dosovitskiy, J.~T. Springenberg, and T.~Brox.
\newblock Learning to generate chairs with convolutional neural networks.
\newblock In {\em CVPR}, 2015.

\bibitem{Erhan09}
D.~Erhan, Y.~Bengio, A.~Courville, and P.~Vincent.
\newblock Visualizing higher-layer features of a deep network.
\newblock Technical report, University of Montreal, 2009.

\bibitem{Girshick14}
R.~Girshick, J.~Donahue, T.~Darrell, and J.~Malik.
\newblock Rich feature hierarchies for accurate object detection and semantic
  segmentation.
\newblock In {\em CVPR}, 2014.

\bibitem{Glasner11}
D.~Glasner, M.~Galun, S.~Alpert, R.~Basri, and G.~Shakhnarovich.
\newblock Viewpoint-aware object detection and pose estimation.
\newblock In {\em ICCV}, 2011.

\bibitem{Hadsell06}
R.~Hadsell, S.~Chopra, and Y.~LeCun.
\newblock Dimensionality reduction by learning an invariant mapping.
\newblock In {\em CVPR}, 2006.

\bibitem{Hartley04}
R.~I. Hartley and A.~Zisserman.
\newblock {\em Multiple View Geometry in Computer Vision}.
\newblock Cambridge University Press, ISBN: 0521540518, second edition, 2004.

\bibitem{jia2014caffe}
Y.~Jia, E.~Shelhamer, J.~Donahue, S.~Karayev, J.~Long, R.~Girshick,
  S.~Guadarrama, and T.~Darrell.
\newblock Caffe: Convolutional architecture for fast feature embedding.
\newblock {\em arXiv preprint arXiv:1408.5093}, 2014.

\bibitem{Karayev14}
S.~Karayev, M.~Trentacoste, H.~Han, A.~Agarwala, T.~Darrell, A.~Hertzmann, and
  H.~Winnem{\"o}ller.
\newblock Recognizing image style.
\newblock In {\em Proc. BMVC.}, 2014.

\bibitem{Krizhevsky12}
A.~Krizhevsky, I.~Sutskever, and G.~E. Hinton.
\newblock {ImageNet} classification with deep convolutional neural networks.
\newblock In {\em NIPS}, 2012.

\bibitem{Kulkarni15}
T.~D. Kulkarni, W.~Whitney, P.~Kohli, and J.~B. Tenenbaum.
\newblock Deep convolutional inverse graphics network.
\newblock {\em arXiv preprint arXiv:1503.03167}, 2015.

\bibitem{Lai11}
K.~Lai, L.~Bo, X.~Ren, and D.~Fox.
\newblock A large-scale hierarchical multi-view {RGB-D} object dataset.
\newblock In {\em Proc. Intl. Conf. on Robotics and Automation}, 2011.

\bibitem{LeCun89}
Y.~LeCun, B.~Boser, J.~Denker, D.~Henderson, R.~Howard, W.~Hubbard, and
  L.~Jackel.
\newblock Backpropagation applied to handwritten zip code recognition.
\newblock {\em Neural Comput.}, 1(4):541--551, 1989.

\bibitem{Lecun04}
Y.~LeCun, F.-J. Huang, and L.~Bottou.
\newblock Learning methods for generic object recognition with invariance to
  pose and lighting.
\newblock In {\em CVPR}, 2004.

\bibitem{leibe2003}
B.~Leibe and B.~Schiele.
\newblock Analyzing appearance and contour based methods for object
  categorization.
\newblock In {\em CVPR}, volume~2, pages II--409. IEEE, 2003.

\bibitem{Mahendran15}
A.~Mahendran and A.~Vedaldi.
\newblock Understanding deep image representations by inverting them.
\newblock In {\em CVPR}, 2015.

\bibitem{Pearson1901}
K.~Pearson.
\newblock On lines and planes of closest fit to systems of points in space.
\newblock {\em Philosophical Magazine}, 2(11):559--572, 1901.

\bibitem{Pepik2012}
B.~Pepik, M.~Stark, P.~Gehler, and B.~Schiele.
\newblock Teaching {3D} geometry to deformable part models.
\newblock In {\em CVPR}, 2012.

\bibitem{Roweis00}
S.~Roweis and L.~Saul.
\newblock Nonlinear dimensionality reduction by locally linear embedding.
\newblock {\em Science}, 290(5500):2323--2326, 2000.

\bibitem{ILSVRCarxiv14}
O.~Russakovsky, J.~Deng, H.~Su, J.~Krause, S.~Satheesh, S.~Ma, Z.~Huang,
  A.~Karpathy, A.~Khosla, M.~Bernstein, A.~C. Berg, and L.~Fei-Fei.
\newblock {ImageNet Large Scale Visual Recognition Challenge}.
\newblock {\em arXiv:1409.0575}, 2014.

\bibitem{Shepard71}
R.~Shepard and J.~Metzler.
\newblock Mental rotation of three dimensional objects.
\newblock {\em Science}, 171(972):701--3, 1971.

\bibitem{Simonyan13}
K.~Simonyan, A.~Vedaldi, and A.~Zisserman.
\newblock Deep inside convolutional networks: Visualising image classification
  models and saliency maps.
\newblock In {\em ICLR workshop}, 2014.

\bibitem{Tanaka93}
K.~Tanaka.
\newblock Neuronal mechanisms of object recognition.
\newblock {\em Science}, 262(5134):685--688, 1993.

\bibitem{Tenenbaum00b}
J.~Tenenbaum, V.~de~Silva, and J.~Langford.
\newblock A global geometric framework for nonlinear dimensionality reduction.
\newblock {\em Science}, 290(550):2319--2323, December 2000.

\bibitem{Turk01}
M.~Turk and A.~Pentland.
\newblock Eigenfaces for recognition.
\newblock {\em J. of Cognitive Neuroscience}, 3(1):71--86, 1991.

\bibitem{Vondrick2013hoggles}
C.~Vondrick, A.~Khosla, T.~Malisiewicz, and A.~Torralba.
\newblock {HOGgles}: Visualizing object detection features.
\newblock In {\em ICCV}, 2013.

\bibitem{Wu15}
Z.~Wu, S.~Song, A.~Khosla, F.~Yu, L.~Zhang, X.~Tang, and J.~Xiao.
\newblock {3D ShapeNets}: A deep representation for volumetric shape modeling.
\newblock In {\em CVPR}, 2015.

\bibitem{Xiang14}
Y.~Xiang, R.~Mottaghi, and S.~Savarese.
\newblock Beyond {PASCAL}: A benchmark for {3D} object detection in the wild.
\newblock In {\em WACV}, 2014.

\bibitem{Yosinski14}
J.~Yosinski, J.~Clune, Y.~Bengio, and H.~Lipson.
\newblock How transferable are features in deep neural networks?
\newblock In {\em NIPS}, 2014.

\bibitem{Zeiler14}
M.~Zeiler and R.~Fergus.
\newblock Visualizing and understanding convolutional networks.
\newblock In {\em ECCV}, 2014.

\bibitem{Zhou15}
B.~Zhou, A.~Khosla, A.~Lapedriza, A.~Oliva, and A.~Torralba.
\newblock Object detectors emerge in deep scene {CNNs}.
\newblock In {\em ICLR}, 2015.

\bibitem{Zhou14}
B.~Zhou, A.~Lapedriza, J.~Xiao, A.~Torralba, and A.~Oliva.
\newblock Learning deep features for scene recognition using {Places} database.
\newblock In {\em NIPS}, 2014.

\end{thebibliography}
}

\end{document}